\documentclass[final,5p,times,twocolumn,authoryear]{elsarticle}

\usepackage{float}
\usepackage{soul}
\usepackage{amsmath,amssymb,amsfonts}
\usepackage{scalerel}
\usepackage{algorithmic}
\usepackage{graphicx}
\usepackage{textcomp}
\usepackage{adjustbox}
\usepackage{xcolor}
\def\BibTeX{{\rm B\kern-.05em{\sc i\kern-.025em b}\kern-.08em
    T\kern-.1667em\lower.7ex\hbox{E}\kern-.125emX}}
\usepackage{subcaption}
\usepackage[colorlinks=true,
            linkcolor=blue,
            citecolor=blue,
            urlcolor=blue]{hyperref}
\usepackage{tabularx}
\usepackage{multirow}
\usepackage{enumitem}
\usepackage{makecell}

\newcommand{\figurewidth}{0.44}

\journal{Neural Networks}

\begin{document}

\begin{frontmatter}



\title{Fully Trainable Deep Differentiable Logic Gate Networks and Lookup Table Networks} %% Article title

\author[imec,vub]{Wout Mommen\corref{cor1}}
\ead{wout.mommen@imec.be}
\cortext[cor1]{Corresponding author.}
\author[imec]{Lars Keuninckx} %% Author name
\ead{lars.keuninckx@imec.be}
\author[imec]{Matthias Hartmann} %% Author name
\ead{matthias.hartmann@imec.be}
\author[imec]{Werner Van Leekwijck} %% Author name
\ead{werner.vanleekwijck@imec.be}
\author[imec,vub]{Piet Wambacq} %% Author name
\ead{piet.wambacq@imec.be}

\affiliation[imec]{organization={imec},%Department and Organization
            addressline={Kapeldreef 75}, 
            city={Leuven},
            postcode={3001}, 
            state={},
            country={Belgium}}
            
\affiliation[vub]{organization={Vrije Universiteit Brussel},%Department and Organization
            addressline={Peinlaan 2}, 
            city={Elsene},
            postcode={1050}, 
            state={},
            country={Belgium}}

%% Abstract
\begin{abstract}
We introduce a novel method for both partial and full optimization of the connections in deep differentiable logic gate networks (LGNs) and lookup table networks (LUTNs). Our training method utilizes a probability distribution over a set of connections per gate/lookup table (LUT) input pin, selecting the connection with highest merit, all whilst the optimal gate types or LUT-entries are learned in parallel. We show that the connection-optimized LGNs outperform standard fixed-connection LGNs on the Yin-Yang, MNIST Handwritten Digits and Fashion-MNIST benchmarks, while requiring only a fraction of the number of logic gates. We achieve 98.92\% on the MNIST dataset with two layers of 8000 gates. With only one layer of 8000 gates, we obtain 98.45\%, showing that our method requires almost 50 times fewer gates compared to fixed-connection LGNs. Training stability up to ten layers has been ensured by employing a high learning rate, straight-through estimators and trimming constant-output gate types. Additionally, we present a LUT neuron description that enables stable training with backpropagation, tested up to 6-layer deep networks. The model requires four times fewer trainable parameters and still achieves a higher accuracy compared to the fixed-connection LGN training algorithm. Our connection-training algorithm also works well for the LUTNs, achieving an accuracy of 98.88\% for two layers of 2000 6-input LUTs.
\end{abstract}


%\item We introduce trainable connections for logic gate networks (LGNs) and lookup table networks (LUTNs).
%\item We demonstrate stable training of LGNs up to 10 layers deep.
%\item Almost 50$\times$ fewer gates are needed to obtain a similar MNIST accuracy.
%\item Stable training of both connections and LUTs is demonstrated.
%\item Fully trainable LGNs (16K gates) and 6-LUTNs (4K 6-LUTs) achieve 98.92\% and 98.88\% on MNIST.

%% Keywords
\end{frontmatter}


\section{Introduction}
In recent years it has become clear that running AI models requires a lot of power \citep{sevilla_compute_2022, desislavov_trends_2023}. As such, a strong interest has been shown to make these models and their associated hardware more energy efficient, especially on the inference side \citep{rastegari_xnornet_2016,hubara_quantized_2018}. One avenue of research is the field of neuromorphic computing that uses bio-inspired Spiking Neural Networks (SNNs) \citep{schuman_opportunities_2022,eshraghian_training_2023}. Here, energy is saved by considering all interactions between the neurons of the network to be binary events. In that way, the multiply-accumulate (MAC) operations that are needed for determining the neuron activations are reduced to accumulate-only operations. These SNNs can be sparsified \citep{rathi_stdpbased_2019} and their weights quantized \citep{putra_qspinn_2021} to a low number of bits, saving energy by using less memory.  Clearly, neuromorphic methods such as SNNs have delivered in their promise for energy efficiency \citep{diehl_truehappiness_2016,kim_spikingyolo_2020}. However the question arises if an even better solution is possible.\\

Recently, it has become clear that directly training a network of Boolean gates, i.e. the very fabric of digital hardware, is possible \citep{petersen_deep_2022}. Deep differentiable logic gate networks (LGNs) are feedforward multi-layered networks of logic gates, in which each gate has two input connections. These networks have fixed randomly chosen connections but trainable gate types. Given two Boolean variables, it is possible to make 16 different 2-input gate types. During training, each gate type is relaxed to a different differentiable expression, such that it can be trained with backpropagation. Apart from the basic feedforward architecture, both a convolutional and recurrent architecture have been realized. The convolutional approach to LGNs showed state-of-the-art performance in throughput speed and accuracy \citep{petersen_convolutional_2024} on the MNIST Handwritten Digits and CIFAR10 data sets. The recurrent architecture for sequence to sequence learning \citep{buhrer_recurrent_2025} showcased a competitive performance compared to conventional deep learning methods on the WMT’14 English-German translation data set. All of these architectures employ non-trainable fixed-connection layers.\\

The concept of LGNs built from two-input gates, can be generalized to multi-input lookup tables (LUTs). LogicNets \citep{umuroglu_logicnets_2020} are quantized sparse neural networks that are converted into networks of LUTs. An X to Y LUT, which is called a logical LUT (L-LUT), is generated from a neuron with X quantized inputs and Y quantized outputs. This L-LUT is converted to 5-to-2 or 6-to-1 LUTs that are physically present in FPGAs. Similarly NullaNet \citep{nazemi_energy-efficient_2019} generates logical LUTs, but converts them to sum-of-product terms. Using a logic synthesis tool, these sum-of-products are then converted to logic gates. PolyLUT \citep{andronic_polylut_2023} also maps neurons to logical X:Y LUTs, but uses multivariate polynomials within the neuron model such that fewer LUTs are needed in the final implementation. NeuraLUT \citep{andronic_neuralut_2024} is the successor to PolyLUT. Instead of using multivariate polynomials, NeuraLUT utilizes an entire sub-network within an L-LUT. This increases its representational capacity, ultimately leading to reduced hardware requirements. They also introduce skip connections in their models, which are hidden within the LUTs. Note that all of these methods do not \textit{directly} train networks of LUTs, instead they \textit{convert} a neural network to a network of LUTs, which might result in more actual physical LUTs required than absolutely necessary. For example, NeuraLUT still uses 54798 LUTs to achieve 96\% on the MNIST Handwritten Digits data set.\\

The following works train networks of lookup tables directly. WARP-LUTs \citep{gerlach_warp-luts_2025} are networks consisting of 2-input LUTs that are trained using the same paradigm as LGNs, but have faster training times and fewer trainable parameters. However, their work is only limited to 2-input LUTs. Differential weightless neural networks (DWNs) \citep{bacellar_differentiable_2024} use an extended finite difference method to train their LUT neuron model, and also learn the interconnect between layers, however no analysis of deep networks with this method is provided. Lookup table networks (LUTNs) \citep{mommen_inter-patient_2026} are similar to DWNs in the sense that they both try to learn a network of LUTs with binary connections using backpropagation. The main difference between the methods is the neuron model: LUTNs train the parameters of the boolean equation of an $2^N$:1 MUX to represent an $N$-input LUT (or $N$-LUT). As such, they use a more interpretable hardware-inspired model to represent the LUT. Note that in these LUTNs, the connections are still fixed during training.\\

In this work we propose a method to extend the paradigm of both LGNs and LUTNs with trainable connections.  
Our contributions are the following:
\begin{enumerate}
    \item We introduce a novel method for learning the connections in LGNs. Currently, the connections in LGNs are chosen at random and stay fixed, hence it is likely not optimal. In this work, the number of candidate connections per gate input pin is a hyperparameter. 
    \item This method is demonstrated on the Yin-Yang, MNIST Handwritten Digits and Fashion-MNIST benchmarks. It is shown that LGNs with partially trained connections require significantly fewer gates, compared to LGNs with fixed connections, for the same accuracy and throughput.
    \item A first attempt at training all possible connections in LGNs with this method is tried out on these three data sets. We achieve a reduction of the network size by almost a factor of 50 (8000 gates instead of 384000 gates) compared to fixed-connection LGNs on the MNIST Handwritten Digits data set, all while obtaining a similar accuracy. With two layers of 8000 gates we obtain an accuracy of 98.92\%. Our training method is stable without loss of accuracy for up to 10 layer deep networks, by employing a higher learning rate, straight-through estimators and by removing constant-output gate types.
    \item We improve the LUTN model by introducing a sigmoid squashing function to keep weights within their respective binary bounds, and perform  annealing to then binarize the trainable LUT parameters. Because of these changes, stable training of deep 6-LUTNs is now possible.
    \item We change the implementation of the LUT-neuron model from element-wise operations to matrix multiplication operations by transforming the calculations to the log-space. This makes the LUTNs more optimized for GPU training, which notably decreases training times.
    \item Compared to the original LGN-training method, we show that our method for training 2-LUTNs obtains a higher accuracy, faster training times, and needs fewer trainable parameters on the MNIST Handwritten Digits benchmark. 
    \item We demonstrate that training the connections of LUTNs with our algorithm significantly increases the accuracy, achieving 98.88\% on a network of two layers of 2000 6-LUTs.
\end{enumerate}

\section{Training method of the LGNs}
\subsection{Standard training method} \label{sec:standard_LGN_training}
The standard way of training the gates in LGNs is by learning a probability distribution over all 16 possible 2-input Boolean operations (i.e. gate types) such as AND, OR and XNOR. A \textit{gate type} is one of the 16 possible 2-input 1-output Boolean functions, as found in \citet{petersen_deep_2022} Table 1. In contrast, we use the term \textit{gate} to describe a single processing unit in our LGN, that after training coincides with one of the 16 aforementioned gate types. During training each single gate in our network is modeled by a probability distribution $(p_g)_{k,i}^l$ over the 16 possible gate types. Here $l$ is the layer number, $k$ is the index of the gate within that layer and $i\in\{0,1,\ldots,15\}$ is the gate-type index. Each Boolean operation is replaced by a differentiable continuous-valued counterpart $f_i$.  The relaxed expressions $f_i$ can be found in \citet{petersen_deep_2022} Table 1. The output value of gate $k$ in layer $l$ is represented in the following way:

\begin{equation} \label{eq:softmax_G}
g_k^l=\sum_{i=0}^{15} (p_g)_{k,i}^l \cdot f_i\left(a_k^{l}, b_k^{l}\right)=\sum_{i=0}^{15} \frac{e^{(w_g)_{k,i}^l/T_g}}{\sum_j e^{(w_g)_{k,j}^l/T_g}} \cdot f_i\left(a_k^l, b_k^l\right).
\end{equation}

Here $a^{l}_k$ and $b^{l}_k$ are the input values to logic gate $k$ of layer $l$. The probability distribution $(p_g)_{k,i}^l$ is given by the output of a softmax function that contains trainable weights $(w_g)_{k,i}^l$. This means that for each gate $k$ in each layer $l$ there is an associated weight for each relaxed gate operation $f_i$. The temperature $T_g$ is a hyperparameter that can be changed during training. After training an LGN, the Boolean operation of the relaxed gate operation $f_i$ matching the highest probability is chosen as in \citet{petersen_deep_2022}. Residual initialization is used to initialize the weight values as described in \citet{petersen_convolutional_2024}, giving a bias to pass-through gates to combat vanishing gradients and promote information flow through the network. Another important concept when working with LGNs, is the discretization step: During training, each gate represents a linear combination of 16 different Boolean operations. After training, we only select one Boolean operation for each gate. Consequently, we observe a drop in accuracy. Luckily, this drop is usually rather small, e.g. for the MNIST Handwritten Digits data set, the drop in accuracy is smaller than 0.1\% \citep{petersen_deep_2022}. This drop in accuracy is related to the discretization gap, which is explained in Section \ref{sec:LGN_experiments} and Section \ref{sec:LUT_experiments}. Apart from the hidden layers that contain gates, we also have counters, also known as group sums. These counters are adopted to predict the correct class label. The binary output values at the end of the network are summed using these counters. If the dimension of the final layer is equal to $D$ and we have $C$ classes, then the population count of the first class is determined by summing the output values of the first $D/C$ gates in the final layer. Here $D$ should be dividable by $C$. The population count of the second class is determined by summing the binary output values of the following $D/C$ output values in the last layer. This is done for all classes, such that each class needs a $\lceil\log_2(D/C+1) \rceil$-bit counter. During training these counter values are passed through a softmax function to convert them to probabilities. Next, they are passed to the loss function and the network can learn the correct gate types using backpropagation.

\subsection{Training the connections of LGNs} \label{sec:LGNs_training_connections}
\begin{figure}[t]
    \centering
     \includegraphics[width=0.4\textwidth,keepaspectratio]{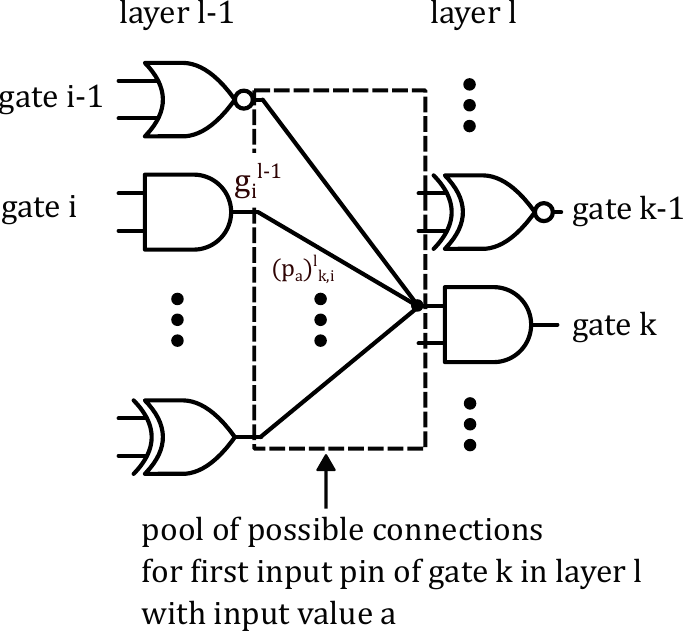}
     \caption{Each first input pin with input value $a$ of gate $k$ in layer $l$ has $N_c$ number of possible connections to gate output pins in the previous layer $l-1$. We will call these the \textit{pool} of possible connections for that input pin with input value $a$ of gate $k$ in layer $l$. Each connection in the pool takes the output value of gate $i$ in the previous layer, namely $g_i^{l-1}$, and multiplies it with a probability value $(p_a)^l_{k,i}$. The sum of these products becomes the actual input value $a$ of gate $k$. The probabilities of all connections in one pool sum to one. The same reasoning holds for the second input value $b$ of gate $k$ in layer $l$. The method of learning connections in this manner is performed in parallel for all inputs, all gates and all layers. At the end of training, for each pool, we select the connection with the highest probability, in essence \textit{binarizing} the connections within the network.}
     \label{fig:drawing_connections}
\end{figure}
We now extend the method of learning the gate operations, to learning the connections to the gates themselves. The algorithm works as follows: For each input pin of each gate and layer, we randomly select $N_c$ different gates from the previous layer. Every single gate input pin in the current layer is then connected to $N_c$ gate output pins in the previous layer. We will call these $N_c$ connections per gate input pin the \textit{pool} of possible connections for that gate input pin. Each connection in a pool of connections is weighted, and these weights are converted to probabilities using a softmax function. During training, these weights will be learned, so the probabilities to the connections in each pool will be learned. At the end of training, we choose the connection with the highest probability in each pool. This is displayed in Figure \ref{fig:drawing_connections}. More formally, the first input value $a$ of gate $k$ in layer $l$ is given by\\
\begin{equation} \label{eq:softmax_C}
a_{k}^{l}=\sum_{i=0}^{N_c-1} (p_a)_{k,i}^{l} \cdot g_{i}^{l-1}=\sum_{i=0}^{N_c-1} \frac{e^{(w_a)_{k,i}^l/T_c}}{\sum_j e^{(w_a)_{k,j}^l/T_c}} \cdot g_{i}^{l-1}.    
\end{equation}

 For the first input pin of each gate $k$ in layer $l$ we randomly choose $N_c$ output pins with value $g_i^{l-1}$ from the previous layer. These output pins are connected to our gate input pin using weighted connections $(w_a)_{k,i}^l$. As mentioned before, the weights are converted to probabilities using a softmax function to obtain $(p_a)_{k,i}^{l}$. To determine the input value $a^l_k$ of the first input pin of gate $k$ in layer $l$, we first multiply the probability value of each connection $(p_a)_{k,i}^{l}$ in the pool with the output value of the gate it is connected to $g_i^{l-1}$. Next, we take the sum of that computation obtaining $a^l_k$. Note that if we want to train all possible connections in a network, $N_c$ must be equal to the number of gates in the previous layer, and this must be true for all layers. The softmax also includes a temperature $T_c$ that determines how uniform or peaked the probability distribution is. At the start of training we use a high $T_c$ (very uniform distribution), which is slowly lowered over the training epochs to a chosen minimum value (very peak distribution). In practice, we observed that a minimum value of $10^{-4}$ was low enough for only selecting one connection for each pool, namely the one with the highest probability. In that way, the softmax is changed into an argmax distribution. For this reason, the softmax function can be seen as a smooth argmax function. The softmax is only calculated across the $i$-dimension (with size $N_c$) of $(w_a)_{k,i}^l$. Since $T_c$ is slowly decreased during training, each vector along the $i$-dimension converges to a one-hot encoded vector. The one-hot encoded vector represents the probability distribution over one pool of possible connections for one gate input pin. This vector is multiplied element-wise with the pool of gate output values from the previous layer. It follows that only the gate output value with the highest probability is selected and used as input value for the gate in the next layer. The expression for the second input value $b_k^l$ is similar, but with probability value $(p_b)_{k,i}^{l}$ and weight matrix $(w_b)_{k,i}^l$.\\ 

 \textbf{Training a subset of connections} In this case all of the connection weights are initialized randomly between 0 and 1 by drawing from a uniform distribution:  $(w_a)_{k,i}^l \sim \mathcal{U}([0,1))$ and $(w_b)_{k,i}^l \sim \mathcal{U}([0,1))$. Using an index tensor, the pool of possible previous layer gate output pins to each current layer gate input pin is determined. By performing element-wise multiplication of this pool of output values with their respective probability values and summing that product, we determined $a_k^l$ and $b_k^l$ (see \eqref{eq:softmax_C}). This happens in parallel for both input values of all gates in each layer. This method is applied in Section \ref{sec:LGN_exp_partial}.\\
 
 \textbf{Training all connections} The LGNs of which we train all possible connections and gates we will call \textit{fully trainable LGNs}. Training all connections, $N_c=all$, is done in a slightly different way. Since we use all possible output pins from the previous layer for each gate input pin, we do not need an index tensor that determines what gate input pin takes into account which output pins from the previous layer. In addition, using high precision indices uses a large amount of memory. Instead, we employ a matrix-vector multiplication of the weights with the input values. This operation is also used in conventional fully connected feedforward neural networks for learning the connection weights. The matrix-vector multiplication replaces the element-wise multiplication and summation operation that was used when training a subset of connections. Furthermore, the weights are initialized differently compared to training a subset of connections. We choose to initialize the connections in a similar way as \citet{keuninckx_training_2026} such that on average, only one connection per gate input pin has a weight value above 0.5. More formally, $(w_a)_{k,i}^l \sim \mathcal{U}([L_s,L_s-1))$ and $(w_b)_{k,i}^l \sim \mathcal{U}([L_s,L_s-1))$, where $L_s$ is the layer size. This method is applied in Section \ref{sec:LGNs_all}.

 \subsection{Straight-through estimator} \label{sec:STE}
 A key element in our method for training fully trainable deep differentiable logic gate networks was the introduction of a straight-through estimator (STE) \citep{kim_deep_2023}. In the forward pass, we will use the argmax function instead of the softmax function. When using an STE on the gates, the forward pass then is represented by
 
\begin{equation} \label{eq:argmax_G}
\begin{aligned}
m_k^l &= \operatorname{argmax_i}[(w_g)_{k,i}^l],\\
g_k^l &= f_{m_k^l}\left(a_k^{l}, b_k^{l}\right).
\end{aligned}
\end{equation}

Here $m_k^l$ is the gate type with the highest weight value for gate $k$ in layer $l$. The remaining symbols follow the convention of \eqref{eq:softmax_G}. The above formula \eqref{eq:argmax_G} we will call the \textit{hard selection} of gates. The backward pass is still treated as if the \textit{soft selection} is performed in the forward pass. This means that the gradients of \eqref{eq:softmax_G} are employed in the backward pass.\\

To use an STE on the connections, we employ the same method. In the forward pass we choose the connection with the highest weight value:

\begin{equation} \label{eq:argmax_C}   
\begin{aligned} 
m_k^l &= \operatorname{argmax_i}[(w_a)_{k,i}^l],\\
a_k^l &= g_{m_k^l}^{l-1}.
\end{aligned}
\end{equation}
Here $m_k^l$ is the connection with the highest weight value in the pool of connections of the first input pin with input value $a$ of gate $k$ in layer $l$. The formula for the second input $b$ is similar, there we have weight matrix $(w_b)_{k,i}^l$. The backward pass is still treated as if we perform the soft selection in the forward pass. Hence, in the backward pass the gradients of \eqref{eq:softmax_C} are still used. The STEs are employed in Section \ref{sec:LGNs_all}, Section \ref{sec:6_LUT_partial} and Section \ref{sec:6_LUT_all}.

\section{Training method of the LUTNs}
\subsection{Standard training method} \label{sec:original_LUTN_method}
The lookup table networks (LUTNs) \citep{mommen_inter-patient_2026} are a generalization of the LGNs; now each ``neuron'' is an $N$-input LUT ($N$-LUT) instead of a 2-input logic gate. These LUTNs pave the way for highly efficient AI applications on field-programmable gate arrays (FPGAs). The reason is that the logic fabric of an FPGA internally makes use of 6-LUTs. This directly leads to highly parallel and low latency implementations. We first train a network of 6-LUTs on a data set, and next convert it to Verilog code which is then mapped to an FPGA, without any additional intermediate steps. The original training method \citep{mommen_inter-patient_2026} makes use of the Boolean equation of a MUX to parametrize the LUT. In that way, it maps naturally onto the FPGA fabric. The method is described in a differentiable form, thus backpropagation can be used for training. To binarize the inputs values of the MUX (i.e. the entries or possible output values of the LUT) a gradual binarization as in \citet{keuninckx_training_2026} was followed. However, the method introduced in \citet{mommen_inter-patient_2026} showed a substantial reduction in accuracy for deep networks, especially for 6-LUTNs. We believe that the main reasons for this decrease in performance are the lack of weight bounding to a small finite range and by the use of a non-differentiable binarization method. We amend this shortcoming by introducing a sigmoid function that limits the outputs of the LUT entries between zero and one, and employ sigmoid-based annealing to binarize the LUT entries. We will now describe this new training method in more detail.\\

\subsection{Sigmoid-based annealing method} \label{sec:sigmoid_annealing}
 The training algorithm has been adapted to use full precision weights that can be both positive and negative during training, and are subsequently run through the operation of a sigmoid function to limit the output values between zero and one. Before the weights are run through the sigmoid, they are scaled by a scalar $\beta$ that starts out small and is increased during training. This effectively amplifies the weight values, such that the output values of the sigmoid become nearly binary. As such, a form of sigmoid annealing is performed. We initialize the weights by drawing from a random uniform distribution between -1 and 1:  $\mathcal{U}([-1,1))$. This limited interval ensures that at the start of training, the gradient of the sigmoid is around its maximum value, boosting learning in the beginning of training. The improved training formula for the LUTs, which is adapted from \citet{mommen_inter-patient_2026}, now reads:

\begin{align}
        L_{out} &= \sum_{i=0}^{2^N-1} \sigma(\beta W_i) \prod_{j=0}^{N-1} \left(\overline{s_{i,j}L_j} +s_{i,j}L_j\right) \label{eq:LUT_eq}\\
        &= \sum_{i=0}^{2^N-1} \sigma(\beta W_i) p(\beta W_i|L_1,\ldots,L_{N-1}) \label{eq:LUT_eq_2} \\
        &= \mathbb{E}[\mathbf{\sigma(\beta W)}], \label{eq:LUT_eq_3}\\
        \text{with} \quad \mathbf{s} &= \begin{pmatrix}
            0 & 0 & \cdots & 0 & 0 \\ 
            0 & 0 & \cdots & 0 & 1 \\ 
            0 & 0 & \cdots & 1 & 0 \\ 
            & & \vdots & & \\
            1 & 1 & \cdots & 1 & 1 \\ \nonumber
        \end{pmatrix}.
    \end{align}
    Here $L_j$ and $\sigma(\beta W_i)$ represent the input values of the LUT and the LUT entries respectively. $L_{out}$ represents the chosen lookup table entry for a certain input pattern. The selection matrix $\mathbf{s}$ has shape $2^N\times N$ and lists all integers from 0 to $2^N-1$ in binary format, where one integer represents one row of the matrix. The matrix selects the input value (or the negation thereof) for each LUT entry. For example, the second entry ($i=1$) of a 3-LUT ($N=3$) is given by $\prod_{j=0}^{N-1} \left(\overline{s_{1,j}L_j} +s_{1,j}L_j\right)=\overline{L_0L_1}L_2$, which are the input values of the second term of the Boolean equation of a MUX \citep{mommen_inter-patient_2026}. The derivation of \eqref{eq:LUT_eq_2} can be found in \ref{sec:Derivation_LUT_Model}.\\

    When the standard binarization method from section \ref{sec:original_LUTN_method} \citep{mommen_inter-patient_2026} was used, the binarization needed to happen layer-wise in a slow fashion over many epochs. By employing the sigmoid-based annealing method, this was not necessary anymore, as all layers of the whole network are annealed at the same time. This reduces the training time significantly as shown in Section \ref{sec:2_LUTs_to_6_LUTs}. To speed up training even more, which is concluded in Section \ref{sec:2_LUTs_to_6_LUTs}, the element-wise (EW) operations were replaced with general matrix multiplications (GEMMs). As a consequence, we moved the calculation to the log-space, and only at the end of the calculation, the exponent of the result was taken:

    \begin{align}
    L_{\text{out}, b, k} &= \exp \left\{ \operatorname{LSE}_{i} \left[ \log p_{b,k,i} + \log \sigma\left(\beta W_{i,k}\right) \right] \right\}, \label{eq:LUT_GEMMs}\\
    \log p_{b,k,i} &= \sum_{j=0}^{N-1} \left( \bar{s}_{i,j} \log \bar{L}_{b,k,j} + s_{i,j} \log L_{b,k,j} \right) \\
    &= \left[\operatorname{log}(\tilde{\mathbf{L}}_{b,k}) \tilde{\mathbf{s}}^T\right]_i. \label{eq:LUT_GEMMs_p}
    \end{align}

    Here $\operatorname{LSE}(x_1,\dots,x_M)=\operatorname{log}(\sum_i\operatorname{exp}(x_i))$ is the LogSumExp function. This LSE function is also known as the smooth max function. In this case, it selects the correct LUT entry based on the received input values to the LUT, but in a smooth and differentiable fashion. This is done in parallel for all LUTs in one layer of the network and for all examples in one batch. The indices $b$ and $k$ refer to the batch index and LUT index respectively. By using concatenation, we get one batched matrix multiplication with dimensions $(B,K,2N)\times(2N,2^N)$ as seen in the last step. Here $B$ is the number of batches, $K$ is the number of LUTs in a layer and $N$ is the number of input pins for a LUT. The derivations of \eqref{eq:LUT_GEMMs} and \eqref{eq:LUT_GEMMs_p} can be found in \ref{sec:Derivation_GEMM}. Unless stated otherwise, the GEMM implementation of the sigmoid-based annealing method is used throughout Section \ref{sec:LUT_experiments}.

    \subsection{Training the connections of LUTNs} \label{sec:training_conn_LUTN}
     The algorithm of section \ref{sec:LGNs_training_connections} for partially and fully training the connections in LGNs was also applied to the LUTNs. Formally, the $m$-th input value $L_m$ of LUT $k$ in layer $l$ is given by:
     \begin{equation} \label{eq:softmax_LUT}
        L_{m,k}^{l}=\sum_{i=0}^{N_c-1} (p_{L_m})_{k,i}^{l} \cdot L_{out,i}^{l-1}=\sum_{i=0}^{N_c-1} \frac{e^{(W_{L_m})_{k,i}^l/T_c}}{\sum_j e^{(W_{L_m})_{k,j}^l/T_c}} \cdot L_{out,i}^{l-1},    
    \end{equation}
with symbol naming conventions as in section \ref{sec:LGNs_training_connections}. The connections of the LUTNs are initialized in the same manner as the connections of the LGNs in Section \ref{sec:LGNs_training_connections}. When partially training the connections of the LUTNs, the weights are initialized as $(W_{L_m})_{k,i}^l \sim \mathcal{U}([0,1))$, while for the full connection training, they are initialized as $(W_{L_m})_{k,i}^l \sim \mathcal{U}([L_s,L_s-1))$, with $L_s$ the layer size. This connection training algorithm is employed in Section \ref{sec:6_LUT_partial} and Section \ref{sec:6_LUT_all}.\\

Having explained our methods, we now move on to describe specific experiments of the LGNs and LUTNs using the proposed algorithms. 

\section{LGN experiments} \label{sec:LGN_experiments}
\subsection{Training a subset of all possible connections of the LGNs} \label{sec:LGN_exp_partial}
\subsubsection{Yin-Yang} \label{sec:YinYang}
\begin{figure}[t]
    \centering
     \includegraphics[width=0.4\textwidth,keepaspectratio]{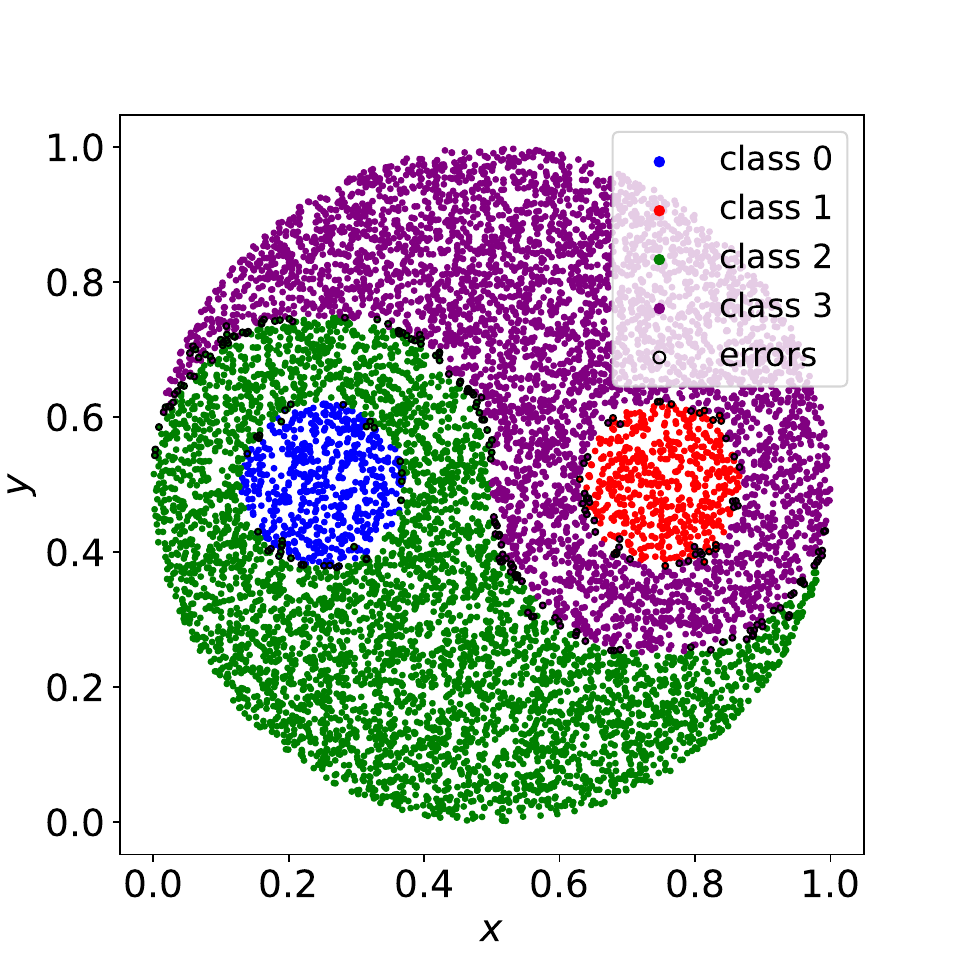}
     \caption{The test set of the Yin-Yang data set using a 2-layer network of 500 gates/layer with the number of trainable connections per gate input pin ($N_c$) equal to 16. A test accuracy of 99.22\% was observed on these 10 000 data points. All errors lie near the edges between the different areas, since the points that are located at the edges are the hardest to classify correctly.}
     \label{fig:YinYang}
\end{figure}

\begin{figure}[t]
     \centering
     \begin{subfigure}[]{\figurewidth\textwidth}
         \centering
         \includegraphics[width=\textwidth]{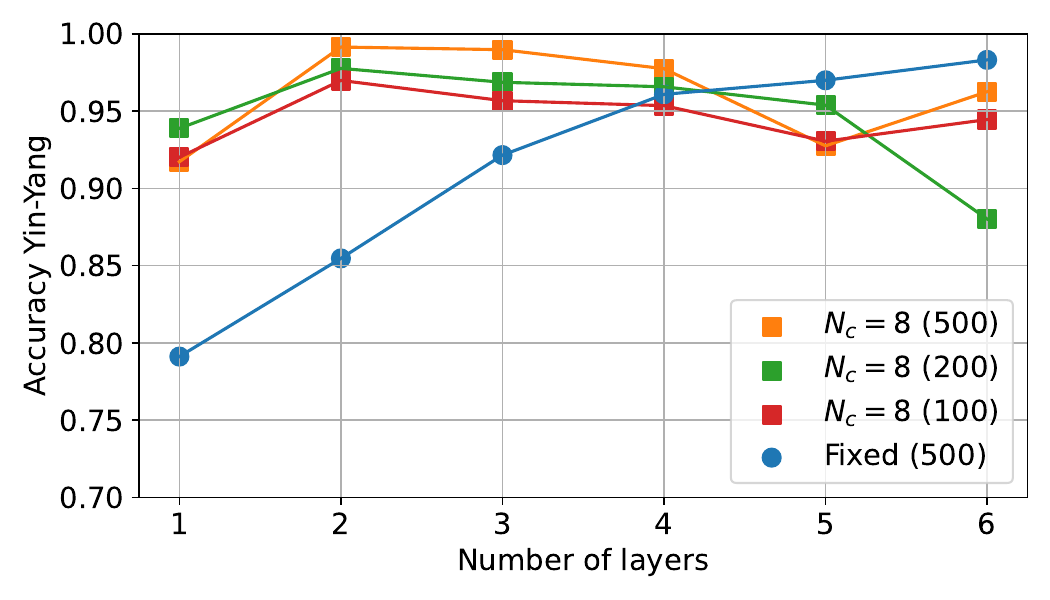}
         \caption{}
         \label{fig:YinYang_Nc8}
     \end{subfigure}
     \hfill
     \begin{subfigure}[]{\figurewidth\textwidth}
         \centering
         \includegraphics[width=\textwidth]{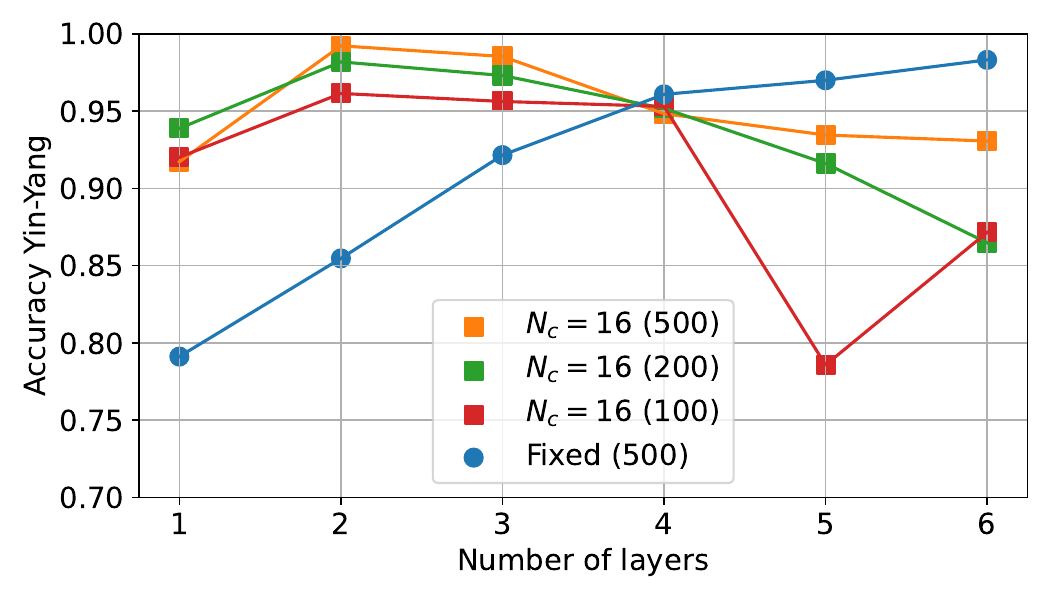}
         \caption{}
         \label{fig:YinYang_Nc16}
     \end{subfigure}
        \caption{Accuracy on the Yin-Yang benchmark as a function of number of layers for (a) $N_c=8$ and (b) $N_c = 16$. Here $N_c$ is the number of trainable connections per gate input pin.  The LGNs with trainable connections are compared to an LGN with fixed connections. The number between brackets indicates the number of gates per layer.}
        \label{fig:YinYang_training}
\end{figure}

The goal of the Yin-Yang data set \citep{kriener_yinyang_2022} is to classify points in the $xy$-plane as belonging to one of four possible regions, as shown in Figure\,\ref{fig:YinYang}. The $x$ and $y$ coordinates are both encoded into 12-bit vectors, such that the input dimension of the network equals 24. The training set consists of 200\,000 data points and the test set contains 10\,000 data points. The accuracy as a function of number of layers can be seen in Figure \ref{fig:YinYang_Nc8}. The number of trainable connections per input pin ($N_c$) in this figure is equal to 8. Figure \ref{fig:YinYang_Nc16} also displays the accuracy as a function of numbers of layers, but here we choose a higher $N_c$ equal to 16. A value of 12 is chosen for $N_c$ for the first layer in both figures. All models were trained with a learning rate of 0.01 for 100 epochs on the Yin-Yang data set. For the models with trainable connections, the softmax temperature of the connections $T_c$ is lowered from 1 to $10^{-4}$ between epochs 60 and 80. Additionally, the softmax temperature of the gates $T_g$ is lowered from 1 to $10^{-4}$ between epochs 80 and 100. From Figures \ref{fig:YinYang_Nc8} and \ref{fig:YinYang_Nc16}, it is clear that training the connections improves the accuracy significantly for shallow networks. All models that are up to three to four layers deep with trainable connections perform better than their fixed connection counterparts. For example, a network with trainable connections consisting of 200 gates (two layers of 100 gates/layer) achieves a higher accuracy (96.98\%) than a network with fixed connections of 2000 gates (four layers of 500 gates/layer, 96.08\%), resulting in one order of magnitude fewer gates and thus also connections. Since the final layer now contains 100 gates instead of 500 gates, we only need four 5-bit counters instead of four 7-bit counters. 

\subsubsection{MNIST Handwritten Digits} \label{sec:LGN_MNIST}
\begin{figure}[t]
     \centering
     \begin{subfigure}[h!]{\figurewidth\textwidth}
         \centering
         \includegraphics[width=\textwidth]{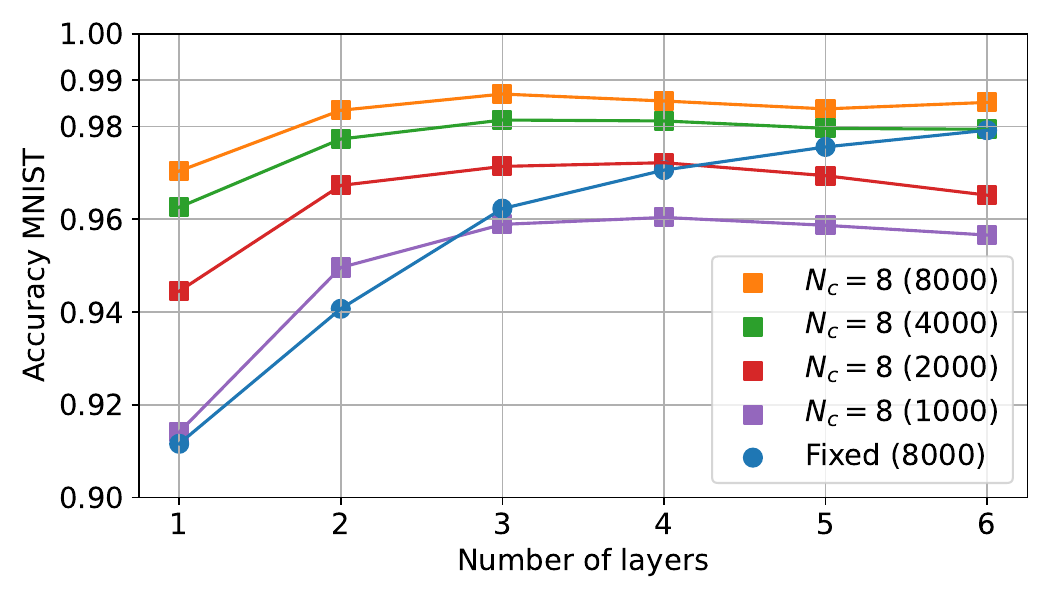}
         \caption{}
         \label{fig:MNIST_Nc8}
     \end{subfigure}
     \hfill
     \begin{subfigure}[h!]{\figurewidth\textwidth}
         \centering
         \includegraphics[width=\textwidth]{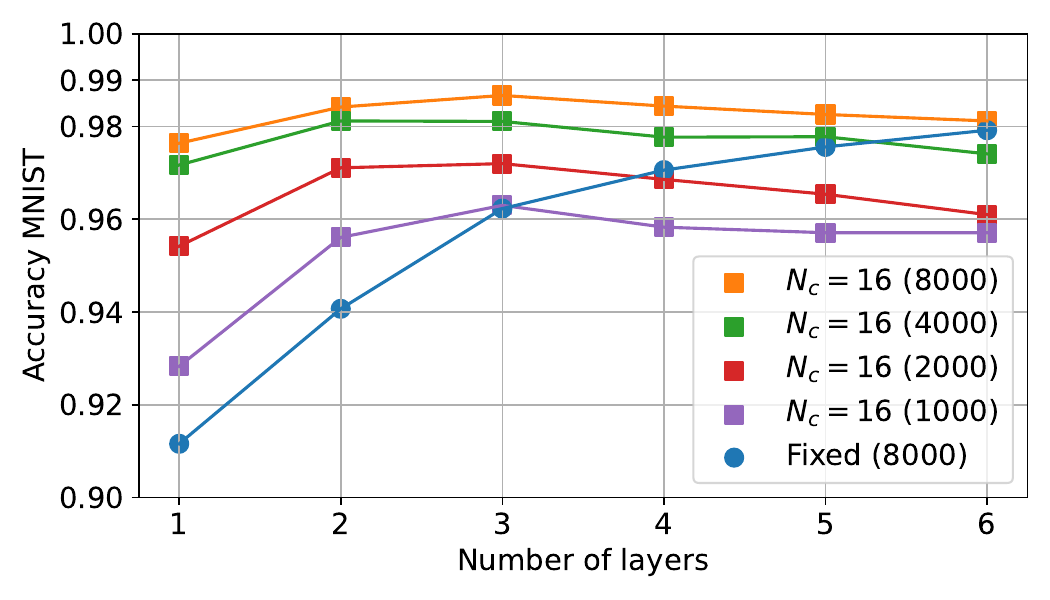}
         \caption{}
         \label{fig:MNIST_Nc16}
     \end{subfigure}
        \caption{Accuracy on the MNIST Handwritten Digits data set as a function of the number of layers for (a) $N_c=8$ and (b) $N_c = 16$ compared with a network of fixed connections. Here $N_c$ is the number of trainable connections per gate input pin. The number between brackets indicates the number of gates per layer.}
        \label{fig:MNIST_training}
\end{figure}

The MNIST Handwritten Digits data set \citep{lecun_mnist_1998} consist of 28x28 grayscale images of handwritten digits ranging from 0 to 9. Before inputting the 784 pixel values $x_i\in[0,1]$ into the model, a threshold of 0.5 is used to binarize them such that $x_i\in\{0,1\}$. Figures \ref{fig:MNIST_Nc8} and \ref{fig:MNIST_Nc16} give the accuracy on the MNIST Handwritten Digits data set as a function of number of layers, for a number of trainable connections per gate input pin of $N_c=8$ and $N_c=16$ respectively. Here, the models were trained with a learning rate of 0.01. The models with fixed or trainable connections were trained for 200 or 240 epochs respectively. In the latter case, the softmax temperature of the connections $T_c$ is lowered from epoch 160 until epoch 200 and the softmax temperature of the gates  $T_g$ is lowered from epoch 200 until epoch 240, using the same start and end values as mentioned in section \ref{sec:YinYang}. The networks with trainable connections again outperform those with fixed connections, for the same number of gates/layer. The largest improvement is found for shallow networks. For example a 3-layer network of 4000 gates/layer with trainable connections performs better (98.14\%) than a 6-layer fixed-connection network of 8000 gates/layer (97.92\%). This is an improvement by a factor of 4 with respect to the number of gates and connections. Another example is the single-layer network of 1000 gates with $N_c=16$, which performs better (92.83\%) than a single-layer network with fixed connections (91.16\%), while having 8 times fewer gates and connections. Additionally, we only need ten 7-bit counters (100 gates in the final layer per class) instead of ten 10-bit counters (800 gates in the final layer per class), for accumulating the binary output values to determine the correct class label.

\subsubsection{Fashion-MNIST}
\begin{figure}[t]
     \centering
     \begin{subfigure}[h!]{\figurewidth\textwidth}
         \centering
         \includegraphics[width=\textwidth]{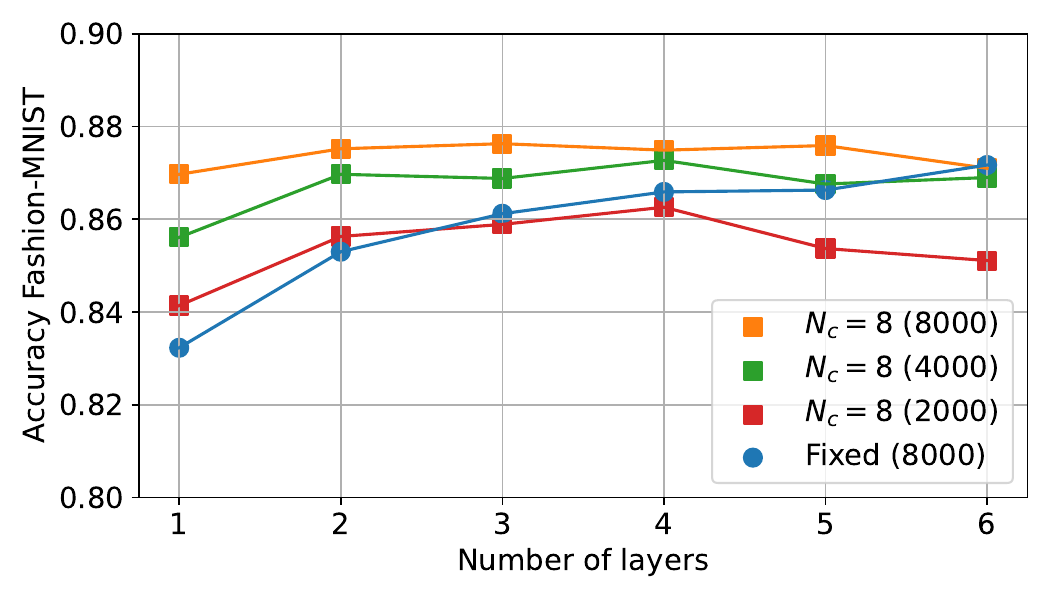}
         \caption{}
         \label{fig:FashionMNIST_Nc8}
     \end{subfigure}
     \hfill
     \begin{subfigure}[h!]{\figurewidth\textwidth}
         \centering
         \includegraphics[width=\textwidth]{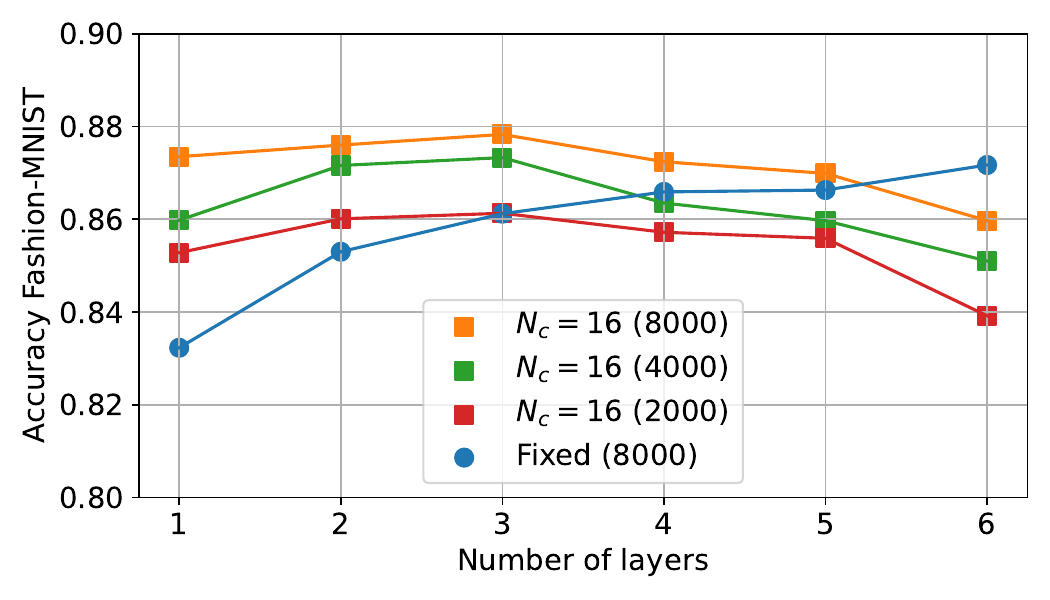}
         \caption{}
         \label{fig:FashionMNIST_Nc16}
     \end{subfigure}
        \caption{Accuracy on the Fashion-MNIST data set as a function of number of layers for (a) $N_c=8$ and (b) $N_c = 16$. Here $N_c$ is the number of trainable connections per gate input pin. These connection-trainable LGNs are compared to their counterpart networks with fixed connections. The number between brackets indicates the number of gates per layer.}
        \label{fig:FashionMNIST_training}
\end{figure}

The Fashion-MNIST data set \citep{xiao_fashionmnist_2017} is similar to the MNIST data set, but instead of handwritten digits, this data set shows 10 classes of clothing items. Thresholds of 0.25, 0.5 and 0.75 are used to encode the 784 grayscale pixel values $x_i\in[0,1]$ into a binary format $x_i\in\{0,1\}$, before applying the data to the network. All LGNs that are trained on the Fashion-MNIST data set are trained using the same settings as for the MNIST Handwritten Digits data set in Section \ref{sec:LGN_MNIST}. The accuracy as a function of number of layers is shown in Figures \ref{fig:FashionMNIST_Nc8} and \ref{fig:FashionMNIST_Nc16} for a given number of trainable connections per gate input pin $N_c = 8$ and $N_c=16$ respectively. Up to five layers, the networks with trainable connections clearly outperform the networks with fixed connections. For example, a network with trainable connections using two layers of 4000 gates/layer has a similar accuracy (87.16\%) as a network using six layers of 8000 gates/layer (87.17\%) with fixed connections, while requiring six times fewer gates and connections. In addition, we only require ten 12-bit counters instead of ten 13-bit counters to determine the class label.

\begin{figure}[h!]
     \centering
     \begin{subfigure}[]{\figurewidth\textwidth}
         \centering
         \includegraphics[width=\textwidth]{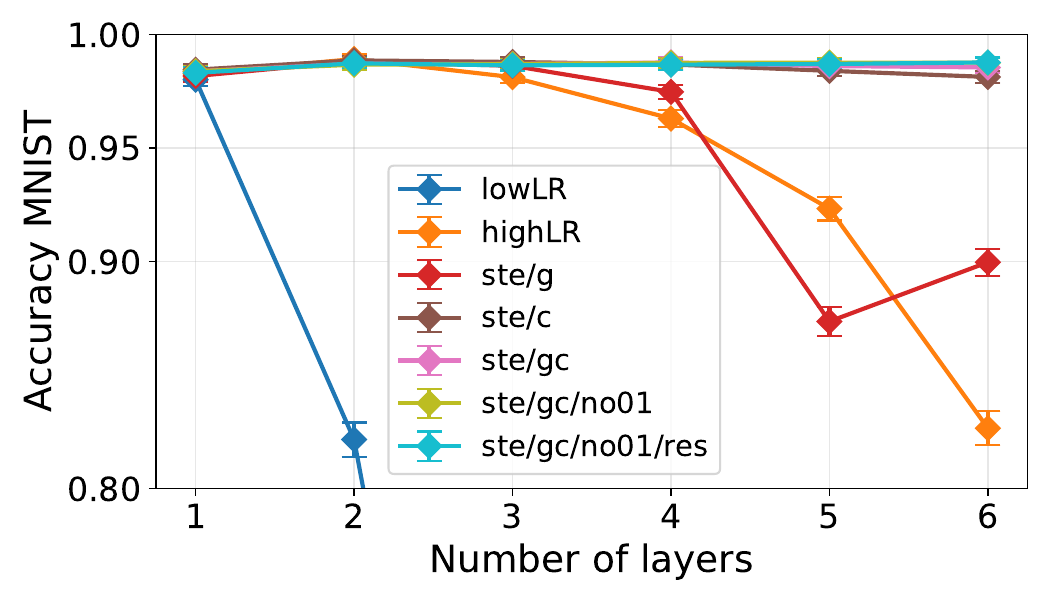}
         \caption{}
         \label{fig:LGN_all_6}
     \end{subfigure}
     \hfill
     \begin{subfigure}[]{\figurewidth\textwidth}
         \centering
         \includegraphics[width=\textwidth]{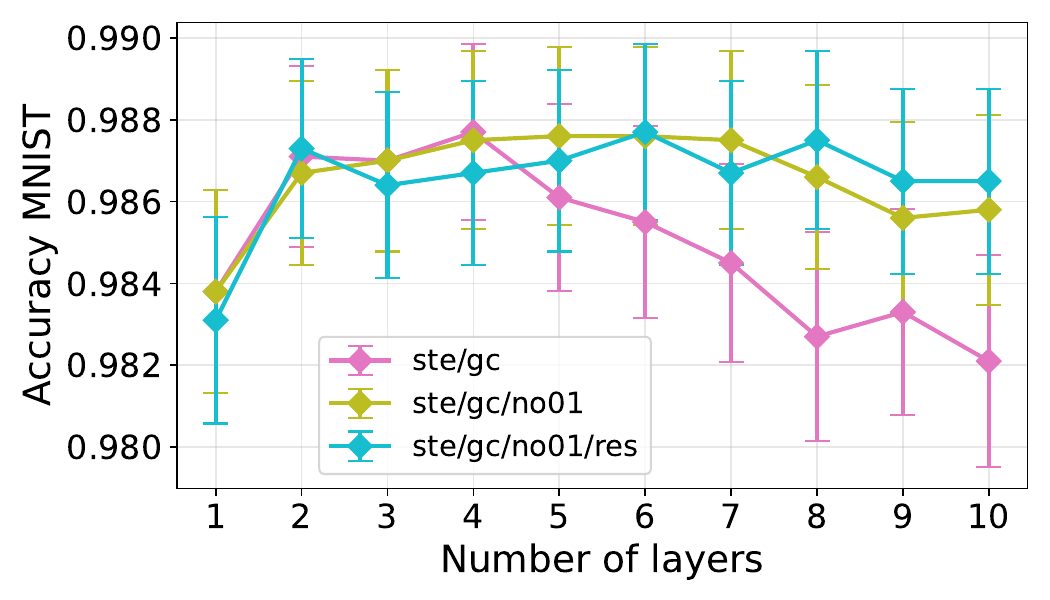}
         \caption{}
         \label{fig:LGN_all_10}
     \end{subfigure}
     \hfill
     \begin{subfigure}[]{\figurewidth\textwidth}
         \centering
         \includegraphics[width=\textwidth]{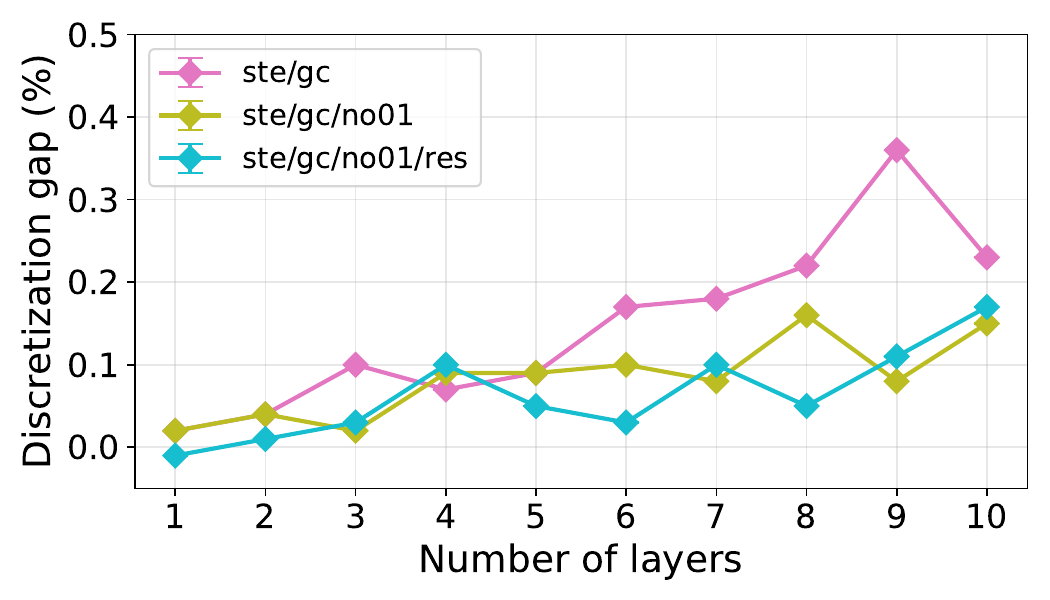}
         \caption{}
         \label{fig:discretization_gap}
     \end{subfigure}
        \caption{Accuracy obtained on the MNIST Handwritten Digits data set as a function of number of layers for a fully trainable LGN, i.e. all of the possible connections and gates are learned as described in Section \ref{sec:LGNs_training_connections}. \textit{lowLR} uses a learning rate of 0.01. All other experiments in this figure use a learning rate of 0.1. Methods labeled with \textit{ste} in their name use a straight-through estimator (STE), namely \textit{ste/g} uses an STE on the gates, \textit{ste/c} uses an STE on the connections and \textit{ste/gc} employs an STE on both gates and connections. The models for which constant-output gates are removed are labeled with \textit{no01}. One model employs residual initialization, which is labeled with \textit{res}. (a) Global view of all methods of Section \ref{sec:LGNs_all} up to six layers. (b) A zoomed version of the accuracy using all methods of Section \ref{sec:LGNs_all} that show a stable training accuracy up to 10 layers. (c) The discretization gap (\%), which is defined as the maximum test set accuracy on the\textit{ full precision} model during training minus the maximum test set accuracy on the \textit{binarized} model during training. With \textit{full precision} we denote the network that still uses softmax expressions for gates \eqref{eq:softmax_G} and connections \eqref{eq:softmax_C}. The term \textit{binarized} is used when we employ the argmax expression for the gates \eqref{eq:argmax_G} and connections \eqref{eq:argmax_C}.}
        \label{fig:LGN_all}
        \vspace{1.3cm}
\end{figure}

\subsection{Training all possible connections of the LGNs} \label{sec:LGNs_all}

We attempt to train all possible connections and gates of LGNs, which we will call \textit{fully trainable LGNs}, for models with different number of layers. Figure \ref{fig:LGN_all_6} gives the accuracy as a function of number of layers on the MNIST Handwritten Digits data set for fully trainable LGNs. A 95\% confidence interval for the accuracy on the test set is utilized for all figures and tables. The 95\% confidence interval on the accuracy $acc$ is given by $[acc-z_{95}\sigma,acc+z_{95}\sigma]$, where $z_{95}\approx1.96$, $\sigma = \sqrt{acc\cdot(1-acc)/n}$ and $n$ is the size of the test set. When training all possible connections, we did not perform any softmax-temperature annealing. By far, the biggest impact on making deep networks fully trainable is the increase of the learning rate. By default, the learning rate of the LGNs is equal to 0.01 (\textit{lowLR}) \citep{petersen_deep_2022}. However, from inspecting Figure \ref{fig:LGN_all_6}, it is seen that the accuracy drops immediately. The accuracies for networks of layers 1-6 while employing this learning rate of 0.01 are equal to 98\%, 82\%, 45\%, 32\%, 16\% and 15\%. As a result, for our experiments, a learning rate of 0.1 (\textit{highLR}) seemed critical when training the network fully. All methods in the figure, except \textit{lowLR}, use a learning rate of 0.1. The LGNs which were trained with method \textit{lowLR} and \textit{highLR} employ \eqref{eq:softmax_G} and \eqref{eq:softmax_C} to train the gates and connections respectively. Yet, even with the high learning rate, it can be noted that for deep networks (i.e. networks with at least four layers) the accuracy still decreases significantly. Inspired by \cite{kim_deep_2023}, we tried out the straight-through estimator (STE) of section \ref{sec:STE} on both the gates and connections. If an STE is implemented, the network has a mechanism to converge more effectively to just one gate type per gate (STE on the gates) and one connection per gate input pin (STE on connections). A gate here is seen as one computational unit within our LGN, while a gate type is one of 16 possible 2-input Boolean operations. When using an STE on the gates, only the gate type with the highest probability is selected for each gate while performing the forward pass \eqref{eq:argmax_G}. In the backward pass however, the gradients are calculated according to the conventional softmax function \eqref{eq:softmax_G}. This seems to help slightly with deep networks (\textit{ste/g}), as the accuracy for a 4-layer network and 6-layer network increases when employing the STE on the gates. Yet, most of the improvement arose from applying an STE to the connections (\textit{ste/c}). In the forward pass, the connection with the highest probability is chosen for each gate input pin using \eqref{eq:argmax_C}. During the backward pass the gradients are determined using \eqref{eq:softmax_C}. Selecting only one connection per gate input pin in the forward pass helps the network converge better to just one connection per gate input pin.\\

Further improvements included applying an STE, as described in Section \ref{sec:STE}, on both the connections and gates (\textit{ste/gc}), removing the constant-output (`0' and `1') gate types (\textit{ste/gc/no01}) and adding residual initialization (\textit{ste/gc/no01/res}). In Figure \ref{fig:LGN_all_10} a more detailed view of the accuracy as a function of network depth is depicted. This clearly shows that all possible connections and gates of LGNs can be trained simultaneously, without any significant loss in accuracy for deep networks up to 10 layers. When the post-training gate distribution was determined, the constant-output gate types occurred most out of any gate type in every layer except the last one. This phenomenon was observed for every network, regardless of network depth. However, constant-output gate types prevent information flow between the layers, as their output values do not depend on their input values. As the amount of information that is dropped by constant-output gate types increases with the number of layers, the accuracy drop for including these gate types is more significant for deeper networks. Hence, as can be seen in Figure \ref{fig:LGN_all_10}, the removal of constant-output gate types `0' and `1' prevents accuracy loss for networks with more than four layers. Another advantage for removing the constant-output gate types is that we only require 14 trainable parameters per gate instead of 16. Only in the last layer we kept constant-output gate types, since they provide trainable biases for the group sums. Additionally, residual initialization as in \citet{petersen_convolutional_2024} is added. This means that the start of training, a weight value of 5 is given to the first pass-through gate type of each gate in the network. The rest of the gate types are assigned a value of 0. The first pass-through gate type has two input pins $a$ and $b$ and one output pin $f$ such that $f=a$, i.e. it can be seen as a residual connection in the network. Alternatively, one could initialize the second pass-through gate type $f=b$ in this manner. Choosing the weight values in this way means that each gate consists for 91\% out of the first pass-through gate type. If we do not take the constant-output gate types into account, such that we only need to apply the softmax to the weight values of 14 gate types, this value increases to 92\%. Adding residual initialization makes sure that even deeper networks will be able to train well \citep{petersen_convolutional_2024}. Figure \ref{fig:LGN_all_10} shows that the current training method is compatible with residual initialization. This is an important property for exploring the training stability of networks with more than 10 layers.\\

Training both the gate types and connections in an LGN means that we can use a much smaller network compared to a fixed-connection LGN while obtaining a similar accuracy. As an example, a single-layer network of 8000 gates achieves 98.45\% on the MNIST Handwritten Digits data set. To compare, a fixed-connection feedforward LGN achieves 98.47\% using 384000 gates \citep{petersen_deep_2022}. As a result, our fully trainable LGN requires close to 50 times fewer gates. A fixed-connection convolutional LGN obtains 98.47\% with 147000 gates on this data set \citep{petersen_convolutional_2024}. Even then, our feedforward network requires nearly 20 times fewer gates. In addition, with two layers of 8000 gates, we manage to obtain an accuracy of 98.92\% on the MNIST Handwritten Digits data set. The accuracy values of 98.45\% for one layer of 8000 gates and 98.92\% for two layers of 8000 gates are not shown in Figure \ref{fig:LGN_all_10} and not mentioned later on in Table \ref{tab:acc_width}, since the former result was obtained with method $ste/c$ and the latter with method $highLR$. Although these methods resulted in a high accuracy for shallow networks, they performed worse on networks beyond two layers deep. For training the connections of deep LGNs, the $ste/gc/no01/res$ method still outperforms the other methods in Figure \ref{fig:LGN_all_6} and Figure \ref{fig:LGN_all_10}.\\

Aside from analyzing the accuracy results of fully trainable LGNs, we also investigated the discretization gap. The discretization gap is defined as the maximum accuracy of the \textit{full precision} model on the test set minus the maximum accuracy of the \textit{binarized} model on the test set during training. By \textit{full precision} we refer to the use of softmax expressions for gates and connections as in \eqref{eq:softmax_G} and \eqref{eq:softmax_C}. In contrast, \textit{binarized} refers to the hard selection of gates \eqref{eq:argmax_G} and connections \eqref{eq:argmax_C}.  In general, with a fixed number of epochs, the best full precision model does not occur at the same epoch as the best binarized model. Note that the discretization gap has been investigated to determine how well each gate converges to one gate type, since the full precision model cannot be implemented in hardware. In Figure \ref{fig:discretization_gap}, the discretization gap is shown for the same methods as in Figure \ref{fig:LGN_all_10}, namely method that use an STE on the gates and connections, that remove constant-output gates and that employ residual initialization. From this figure, we see that removing constant-output gate types (`0' and `1') reduces the discretization gap significantly, which increases the accuracy of the binarized model. The discretization gap increases only slightly with depth for the models without constant gates. The use of residual initialization does not affect the discretization gap for networks up to 10 layers. For the remainder of this section, we use a learning rate of 0.1, an STE on gates and connections, removal of constant-output gates, and residual initialization as a default training settings for all the experiments.\\

\begin{table*}[t]
\centering
\caption{Maximum test accuracy (\%) of the fully trainable LGNs using a straight-through estimator (STE) on both the gates and connections. Constant-output gates are removed in all layers except the final one before training. Additionally, residual initialization is used to give a bias to pass-through gate types. The accuracy is determined on the Yin-Yang, MNIST Handwritten Digits, and Fashion-MNIST benchmarks for varying number of layers and gates/layer.}
\label{tab:acc_width}
\setlength{\tabcolsep}{4pt}
\begin{tabular}{c|ccc|cccc|ccc}
\hline
 & \multicolumn{3}{c|}{Yin-Yang} & \multicolumn{4}{c|}{MNIST Handwritten Digits} & \multicolumn{3}{c}{Fashion-MNIST} \\
Layers & 100 & 200 & 500 & 1k & 2k & 4k & 8k & 2k & 4k & 8k \\
\hline
1 & $91.0\pm0.6$ & $93.6\pm0.5$ & $94.4\pm0.4$ & $95.6\pm0.4$ & $97.0\pm0.3$ & $97.9\pm0.3$ & $98.3\pm0.3$ & $86.7\pm0.7$ & $87.4\pm0.7$ & $88.3\pm0.6$ \\
2 & $91.3\pm0.6$ & $95.9\pm0.4$ & $96.8\pm0.3$ & $96.3\pm0.4$ & $97.3\pm0.3$ & $98.1\pm0.3$ & $98.7\pm0.2$ & $87.1\pm0.7$ & $87.7\pm0.6$ & $88.4\pm0.6$ \\
3 & $91.5\pm0.5$ & $97.3\pm0.3$ & $98.1\pm0.3$ & $96.6\pm0.4$ & $97.4\pm0.3$ & $98.4\pm0.2$ & $98.6\pm0.2$ & $87.2\pm0.7$ & $87.8\pm0.6$ & $88.4\pm0.6$ \\
4 & $92.0\pm0.5$ & $97.6\pm0.3$ & $98.4\pm0.2$ & $96.5\pm0.4$ & $97.3\pm0.3$ & $98.3\pm0.3$ & $98.7\pm0.2$ & $86.9\pm0.7$ & $87.9\pm0.6$ & $88.2\pm0.6$ \\
5 & $91.9\pm0.5$ & $97.7\pm0.3$ & $98.7\pm0.2$ & $96.6\pm0.4$ & $97.3\pm0.3$ & $98.2\pm0.3$ & $98.7\pm0.2$ & $86.8\pm0.7$ & $87.7\pm0.6$ & $88.1\pm0.6$ \\
6 & $93.7\pm0.5$ & $97.7\pm0.3$ & $98.6\pm0.2$ & $96.5\pm0.4$ & $97.1\pm0.3$ & $98.1\pm0.3$ & $98.8\pm0.2$ & $86.6\pm0.7$ & $87.4\pm0.7$ & $88.1\pm0.6$ \\
\hline
\end{tabular}
\end{table*}

\begin{table}[t]
\centering
\caption{Discretization gap (\%) for varying number of layers and gates/layer. The discretization gap is defined as the maximum test set accuracy on the \textit{full precision} model during training minus the maximum test set accuracy on the \textit{binarized} model during training, even if they do not occur on the same epoch. With the term \textit{full precision} we denote the network that still uses softmax expressions for gates \eqref{eq:softmax_G} and connections \eqref{eq:softmax_C}. The term \textit{binarized} is used when we employ the argmax expression for the gates \eqref{eq:argmax_G} and connections \eqref{eq:argmax_C}. This binarized model is the \textit{true} binary network as could be implemented in digital hardware. The fully trainable LGNs employed a straight-through estimator (STE) on the connections and gates, used residual initialization, and did not contain constant-output gates in all hidden layers, except the last one.}
\label{tab:gap_width}
\setlength{\tabcolsep}{4pt}
\begin{tabular}{c|ccc|cccc|ccc}
\hline
 & \multicolumn{3}{c|}{Yin-Yang} & \multicolumn{4}{c|}{MNIST Handw. D.} & \multicolumn{3}{c}{Fashion-MNIST} \\
Layers & 100 & 200 & 500 & 1k & 2k & 4k & 8k & 2k & 4k & 8k \\
\hline
1 & $2.0$ & $0.9$ & $0.5$ & $0.7$ & $0.3$ & $0.1$ & $0.0$ & $0.0$ & $0.0$ & $0.1$ \\
2 & $0.0$ & $1.2$ & $1.5$ & $1.0$ & $0.5$ & $0.1$ & $0.0$ & $0.1$ & $0.0$ & $0.1$ \\
3 & $1.3$ & $0.6$ & $0.5$ & $0.9$ & $0.5$ & $0.1$ & $0.0$ & $0.0$ & $0.2$ & $0.1$ \\
4 & $0.4$ & $0.7$ & $0.5$ & $0.8$ & $0.6$ & $0.2$ & $0.1$ & $0.3$ & $0.1$ & $0.0$ \\
5 & $0.6$ & $0.6$ & $0.3$ & $0.8$ & $0.5$ & $0.2$ & $0.0$ & $0.4$ & $0.0$ & $0.0$ \\
6 & $0.5$ & $0.5$ & $0.2$ & $0.9$ & $0.8$ & $0.3$ & $0.0$ & $0.4$ & $0.1$ & $0.2$ \\
\hline
\end{tabular}
\end{table}

Table \ref{tab:acc_width} summarizes the results for the fully trainable LGNs of which the method can be found in Section \ref{sec:LGNs_training_connections}. The table gives the accuracy of the fully trainable LGNs on the Yin-Yang, MNIST Handwritten Digits and Fashion-MNIST data sets for different number of layers and gates/layer. For the Yin-Yang data set, independent of width, the accuracy increases with depth outside of the confidence intervals. For the other two data sets, most of the accuracy increase is made in the first two layers. For networks with more than three layers, the accuracy saturates within the confidence intervals. Across all data sets and network depths, it is noticeable that the wider the network, the better the performance. The reason as to why the network width plays such a major role in increasing the accuracy of the model compared to the network depth is currently unknown. For fixed-connection LGNs on the MNIST Handwritten Digits benchmark, we see an increase in accuracy up to a certain depth, after which the accuracy saturates. Since we now also train the connections, we reach this plateau earlier.\\

The discretization gap for different benchmarks, number of layers and number of gates per layer is given in Table \ref{tab:gap_width}. For the Yin-Yang benchmark, the gap decreases slightly with an increasing number of layers. For both MNIST Handwritten Digits and Fashion-MNIST, it is observed that the gap either increases slightly or remains constant if the network is deeper. However, the most noteworthy trend across all data sets and network depths is that the gap decreases if the number of gates/layer increases. This means that the full-precision network converges better to its binary version when wider layers are used.

\section{LUTN experiments} \label{sec:LUT_experiments}
\subsection{Training the LUTs of LUTNs}
\subsubsection{Comparison with LGNs} \label{sec:LUTs_vs_LGNs}
\begin{figure}[h!]
     \centering
     \begin{subfigure}[]{\figurewidth\textwidth}
         \centering
         \includegraphics[width=\textwidth]{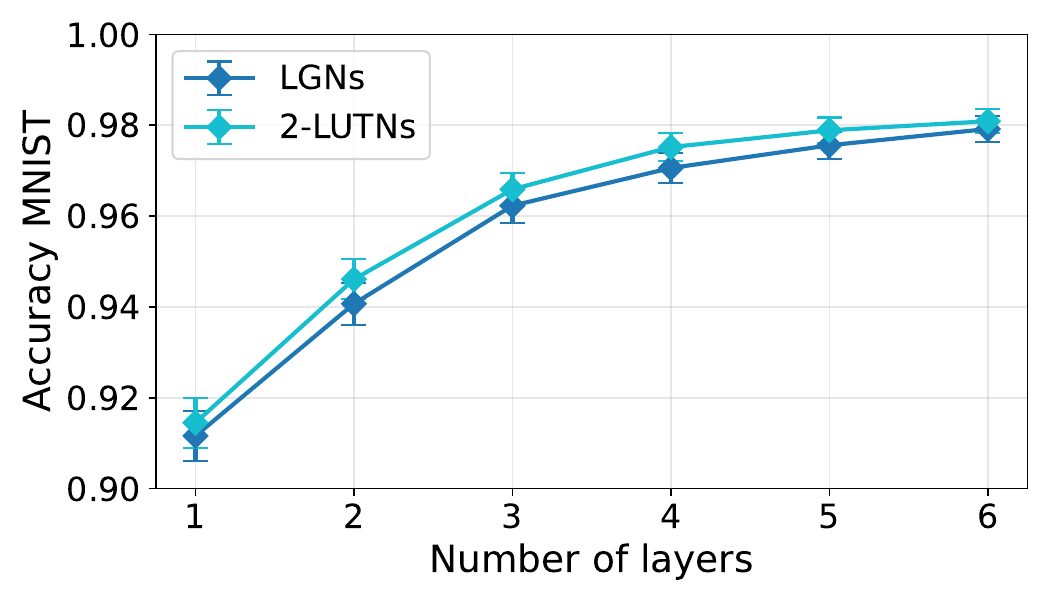}
         \caption{}
         \label{fig:acc_2LUTs}
     \end{subfigure}
     \hfill
     \begin{subfigure}[]{\figurewidth\textwidth}
         \centering
         \includegraphics[width=\textwidth]{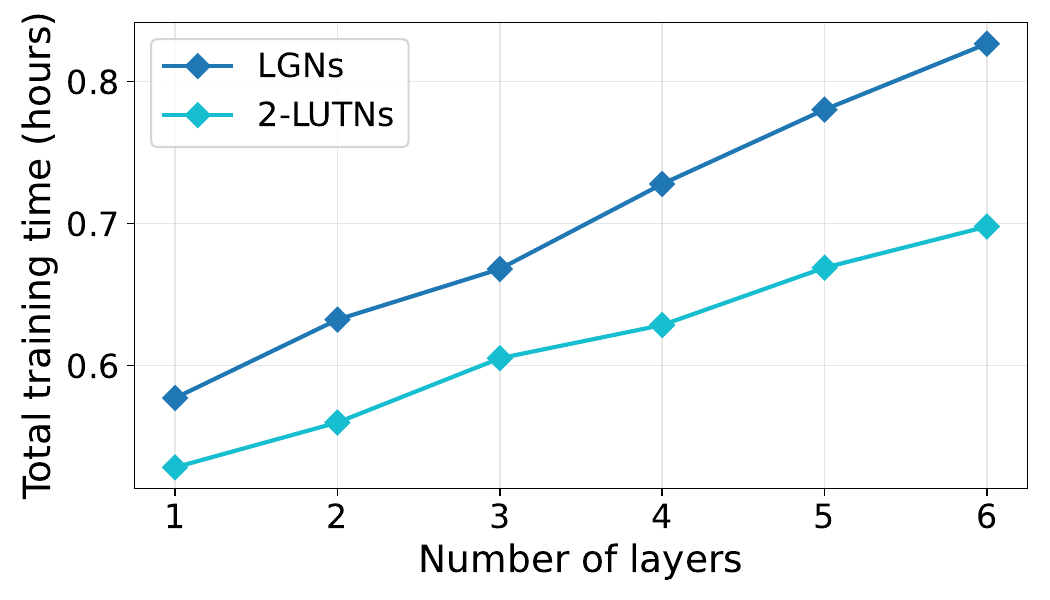}
         \caption{}
         \label{fig:training_time_2LUTs}
     \end{subfigure}
        \caption{(a) Accuracy and (b) training time of the 2-LUTNs (cyan) compared to fixed-connection LGNs (blue), as explained in Section \ref{sec:sigmoid_annealing} and Section \ref{sec:standard_LGN_training} respectively. The 2-LUTNs and LGNs were benchmarked on the MNIST Handwritten Digits data set using an NVIDIA L40S GPU. Both types of networks were given randomly chosen and fixed connections.}
        \label{fig:LUTNs_vs_LGNs}
\end{figure}

First, the fixed-connection LUTNs with 2-LUT units are compared with the fixed-connection LGNs on the MNIST Handwritten Digits data set in terms of accuracy in Figure \ref{fig:acc_2LUTs} and training time in Figure \ref{fig:training_time_2LUTs}. The 2-LUTNs perform better in both accuracy and training time, while having four times fewer parameters compared to fixed-connection LGNs of Section \ref{sec:standard_LGN_training}. This can be seen by comparing \eqref{eq:softmax_G} with \eqref{eq:LUT_eq}. In the former equation, we require 16 trainable parameters per gate, while in the latter equation we only have four trainable parameters $W_i$ for each 2-LUT. Here, we trained all the LUTNs for 200 epochs with a learning rate of 0.01. In all LUTN experiments sigmoid-annealing, as explained in Section \ref{sec:sigmoid_annealing}, is performed from epoch 100 until 200.

\subsubsection{Replacing 2-LUTs with 6-LUTs} \label{sec:2_LUTs_to_6_LUTs}
\begin{figure}[h!]
     \centering
     \begin{subfigure}[]{\figurewidth\textwidth}
         \centering
         \includegraphics[width=\textwidth]{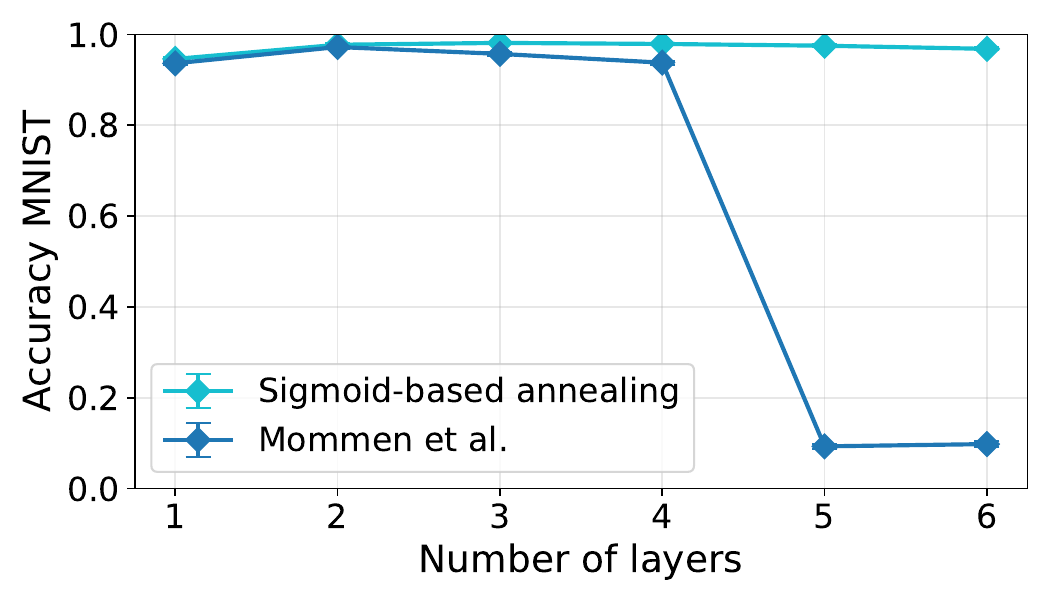}
         \caption{}
         \label{fig:acc_6LUTs_old_vs_new}
     \end{subfigure}
     \hfill
     \begin{subfigure}[]{\figurewidth\textwidth}
         \centering
         \includegraphics[width=\textwidth]{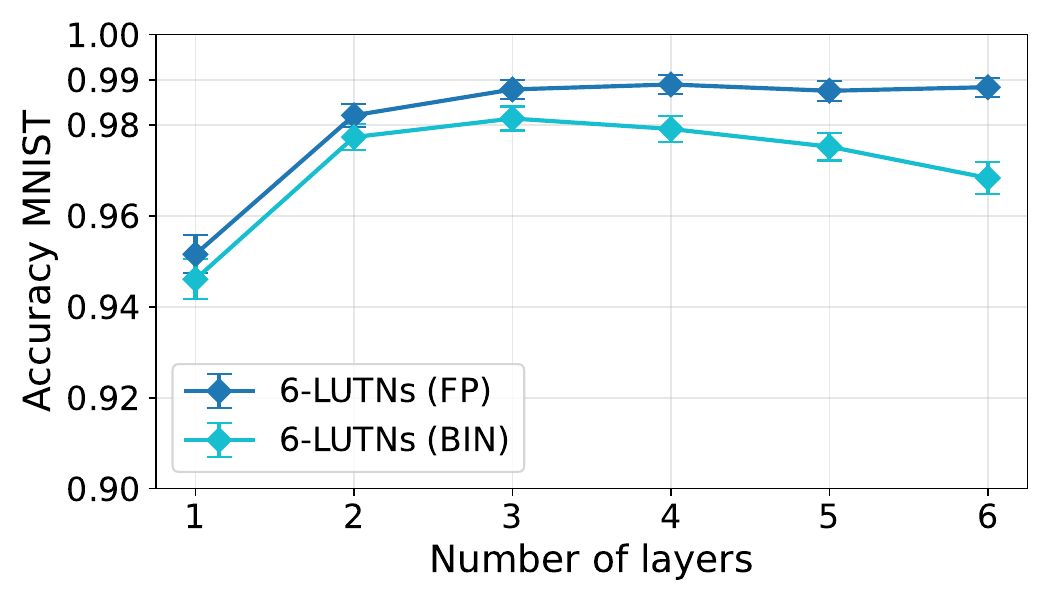}
         \caption{}
         \label{fig:acc_6LUTs}
     \end{subfigure}
        \caption{(a) Comparison of accuracies obtained from the method of \citet{mommen_inter-patient_2026} (blue) and the sigmoid-based annealing method (cyan) from Section \ref{sec:sigmoid_annealing} on the MNIST Handwritten Digits data set. (b) The accuracy on the same data set using the sigmoid-based annealing method. The model with non-binarized LUT entries (FP, blue) is compared with the model containing binarized LUT entries (BIN, cyan).}
        \label{fig:acc_6LUTs_both}
\end{figure}

The algorithm defined by \eqref{eq:LUT_GEMMs} in section \ref{sec:sigmoid_annealing} is compared to the algorithm from \citet{mommen_inter-patient_2026}, Section \ref{sec:original_LUTN_method}. In  Figure \ref{fig:acc_6LUTs_old_vs_new}, the accuracy of both methods on the MNIST Handwritten Digits data set is shown. Here, all of the networks contain 2000 6-LUTs per layer. For networks consisting of at least five layers, the training method of \citet{mommen_inter-patient_2026} does not give meaningful results. In contrast, the sigmoid-based annealing method manages to provide a high accuracy for networks up to six layers. The only caveat of the current sigmoid-based annealing algorithm is the increasing discretization gap for deeper networks, which is shown in Figure \ref{fig:acc_6LUTs}. In this figure, the networks with \textit{full precision} LUT entries show a stable accuracy value from three to six layers. Conversely, the accuracy of the models for which the LUT entries are \textit{binarized} using sigmoid annealing, decreases for a network of three layers to a network of six layers. This behavior is different from LUTNs containing 2-LUTs as described in section \ref{sec:LUTs_vs_LGNs}. In Figure \ref{fig:acc_2LUTs}, the accuracy of the binarized 2-LUTN did not decrease for networks with three to six layers. This means that increasing the number of LUT input pins is the main cause for the increase of the discretization gap. It is important to note that the network with full precision values is a step in between of obtaining the fully binarized LUTN, which is actually implementable in hardware. The goal of this comparison between full precision and binarized network is to help determine the origin of shortcomings of the proposed training algorithm. Using an STE, either from the start, in the middle, or at the end training did not remove the discretization gap. Additionally, employing a higher learning rate which was used for the fully trainable LGNs, degraded performance. Although the accuracy decreases slightly with deep 6-LUTNs, it must be noted that for shallow networks the accuracies of the 6-LUTNs are much higher than the 2-LUTNs. For the 6-LUTNs we observe accuracies of 94.6\%, 97.7\% and 98.2\% for networks consisting of one, two and three layers. In comparison, the 2-LUTNs achieve 91.5\%, 94.6\% and 96.6\% for these network depths. This is especially noteworthy since we only used 2000 6-LUTs/layer instead of the 8000 2-LUTs/layer. The total number of connections to one 6-LUT layer is equal to 12000, while we have 16000 connections for one 2-LUT-layer. Thus, even with fewer connections, the 6-LUT layers perform better than the 2-LUT layers. This significant performance increase might stem from the complexity of the LUTs. An $N$-input LUT can learn one of $2^{2^N}$ binary functions \citep{mommen_inter-patient_2026}. Hence, for a 2-LUT this is equal to only 16 binary functions, while a 6-LUT can learn one out of $1.845\cdot10^{19}$ binary functions.\\

Lastly, we compare the training time durations of the element-wise and GEMM implementation of the sigmoid-based annealing method described in Section \ref{sec:sigmoid_annealing} with method of \citet{mommen_inter-patient_2026} in Table \ref{tab:training_time_speedup_vs_gemm}. The speedup is determined by taking the fraction of training time durations of both methods. From Table \ref{tab:training_time_speedup_vs_gemm} we observe that the GEMM implementation is significantly faster than the element-wise implementation. This is especially the case for deeper networks, i.e. the speedup monotonically increases from a 1-layer network ($\times1.26$) to a 6-layer network ($\times3.60$). Comparing the GEMM implementation to the method of \citet{mommen_inter-patient_2026}, we see that the speedup is quite significant, e.g. up to one order of magnitude for networks with at least four layers.

\begin{table}[ht]
\centering
\begin{tabular}{c|c|c}
\hline
Layer & EW/GEMM & Mommen et al./GEMM \\
\hline
1 & $\times 1.26$ & $\times 1.12$ \\
2 & $\times 1.89$ & $\times 3.32$ \\
3 & $\times 2.45$ & $\times 6.23$ \\
4 & $\times 2.90$ & $\times 9.37$ \\
5 & $\times 3.28$ & $\times 13.62$ \\
6 & $\times 3.60$ & $\times 9.98$ \\
\hline
\end{tabular}
\caption{The second column shows the speedup in training time of the GEMM implementation of the sigmoid-based annealing method compared to the element-wise (EW) implementation. The speedup is determined by dividing the training time of the EW method by the training time of the GEMM method. The third column gives the speedup of the GEMM implementation compared to the method of \citet{mommen_inter-patient_2026}, using the same formula, but replacing the numerator with the training times of the method of \citet{mommen_inter-patient_2026}.}
\label{tab:training_time_speedup_vs_gemm}
\end{table}

\subsection{Training a subset of all possible connections of the LUTNs} \label{sec:6_LUT_partial}
\begin{figure}[t]
    \centering
     \includegraphics[width=0.49\textwidth,keepaspectratio]{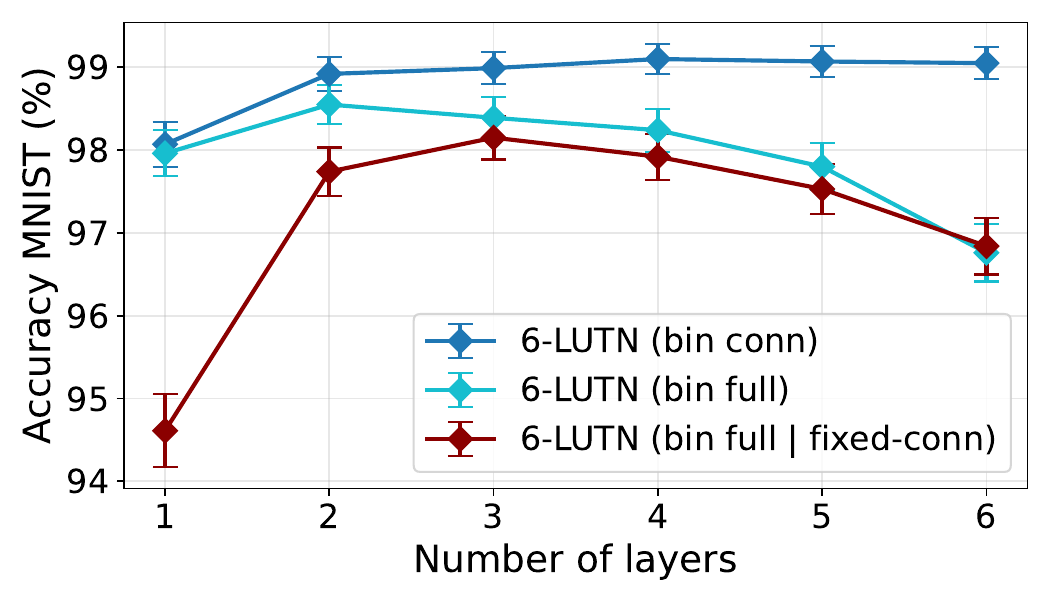}
     \caption{Accuracy on the MNIST Handwritten Digits data set of the 6-LUTN with binarized connections and non-binarized LUTs (blue) and the 6-LUTN that is fully binarized (both connections and LUTs, cyan) for $N_c=8$. The fully binarized 6-LUTN with fixed connections throughout training is also given as a reference (dark red).}
     \label{fig:LUTNs_partial_connections}
\end{figure}

In this section we will train both the LUTs and a subset of all possible connections of the LUTNs. The LUT entries are full precision values during training and are slowly binarized as outlined in Section \ref{sec:sigmoid_annealing}. With \textit{binarized LUTs} we denote that the entries in the LUTs are binarized. Additionally, the connections are trained using the same algorithm \eqref{eq:softmax_LUT} as for the LGNs. Again, the connection with the highest probability is chosen for each pool of possible connections for each LUT input pin at the end of training. Since this is equivalent to changing the softmax of the weights $(W_{L_m})_{k,i}^l$ over dimension $i$ of \eqref{eq:softmax_LUT} to a one-hot encoded vector over that dimension, we will call the LUTN with chosen connections a LUTN with \textit{binarized connections}. Figure \ref{fig:LUTNs_partial_connections} displays the accuracy achieved on the MNIST Handwritten Digits data set of a 6-LUTN with 2000 LUTs/layer and 8 trainable connections per LUT input pin ($N_c=8$). When only the connections are binarized, the performance of the 6-LUTN can get above 99\% after only three layers. To obtain these results, a softmax annealing on the connections is performed first from epoch 75 until epoch 125. Afterwards, sigmoid annealing of the LUTs is executed from epoch 125 until epoch 200, which is also the end of training. When the LUT entries are converted from full precision to binary values, the performance drops significantly. This is shown in Figure \ref{fig:LUTNs_partial_connections}, where from two layers onwards, the gap between the LUTN with only binarized connections and a fully binarized LUTN increases with the number of layers. The algorithm for training the connections in LUTNs also uses an STE in the same way as the LGNs, since it improved the performance. Increasing the learning rate enhanced the performance of the LGNs, but did not enhance the performance of the LUTNs. Although the discretization gap increases with network depth, it is noticeable that the networks perform significantly better than the fixed-connections LUTNs. For one layer of 2000 LUTs we obtain an accuracy of 98.0\% with 8 trainable connections per gate input pin instead of 94.6\% with a fixed-connection LUTN.

\begin{table*}[t]
\centering
\caption{Test set accuracy (\%) versus network depth on the MNIST Handwritten Digits data set for a 6-LUTN of 2000 LUTs/layer. Different number of trainable connections per LUT input pin $N_c$ were chosen for the method of Section \ref{sec:training_conn_LUTN}. The 6-LUTN network in the left part of the table has binarized connections and non-binarized LUTs. The right side has both binarized connections and binarized LUT entries.}
\label{tab:acc_partial_full_nc}
\setlength{\tabcolsep}{4pt}
\begin{tabular}{c|ccc|ccc}
\hline
 & \multicolumn{3}{c|}{Binarized connections \& non-binarized LUTs} & \multicolumn{3}{c}{Binarized connections \& binarized LUTs} \\
Layers & $N_c=8$ & $N_c=16$ & $N_c=\mathrm{all}$ & $N_c=8$ & $N_c=16$ & $N_c=\mathrm{all}$ \\
\hline
1 & $98.1 \pm 0.3$ & $98.4 \pm 0.2$ & $98.6 \pm 0.2$ & $98.0 \pm 0.3$ & $98.3 \pm 0.3$ & $98.5 \pm 0.2$ \\
2 & $98.9 \pm 0.2$ & $99.0 \pm 0.2$ & $99.1 \pm 0.2$ & $98.6 \pm 0.2$ & $98.6 \pm 0.2$ & $98.9 \pm 0.2$ \\
3 & $99.0 \pm 0.2$ & $99.1 \pm 0.2$ & $99.2 \pm 0.2$ & $98.4 \pm 0.2$ & $98.6 \pm 0.2$ & $98.4 \pm 0.2$ \\
4 & $99.1 \pm 0.2$ & $99.1 \pm 0.2$ & $99.2 \pm 0.2$ & $98.2 \pm 0.3$ & $98.3 \pm 0.3$ & $97.9 \pm 0.3$ \\
5 & $99.1 \pm 0.2$ & $99.2 \pm 0.2$ & $99.1 \pm 0.2$ & $97.8 \pm 0.3$ & $97.8 \pm 0.3$ & $95.9 \pm 0.4$ \\
6 & $99.1 \pm 0.2$ & $99.1 \pm 0.2$ & $99.2 \pm 0.2$ & $96.8 \pm 0.3$ & $96.6 \pm 0.4$ & $93.7 \pm 0.5$ \\
\hline
\end{tabular}
\end{table*}

\subsection{Training all possible connections of the LUTNs} \label{sec:6_LUT_all}
We trained both the full set of connections and LUT entries in LUTNs with the algorithm as described in Section \ref{sec:training_conn_LUTN} for the connections and Section \ref{sec:sigmoid_annealing} for the LUT entries. In Table \ref{tab:acc_partial_full_nc} the accuracy on the MNIST Handwritten Digits data set is given for 6-LUTNs with 2000 LUTs per layer. Apart from training every possible connection, we set the number of trainable connections per LUT input pin $N_c$ equal to 8 and 16. If we look at the LUTNs with binarized connections and non-binarized LUTs, we see that performance increases for a higher number of possible connections per gate input pin $N_c$. With only two layers of 2000 LUTs, we achieve an accuracy of 99.1 \%. This result remains stable for networks with more layers. Binarizing the LUTs leads to a noticeable reduction in accuracy, particularly for deeper networks and larger $N_c$ values. For example, the accuracy decreases by only 0.1\% for all single-layer networks, whereas a 6-layer network with $N_c=$ all experiences a 5.5\% drop.

\section{Conclusion}
\begin{table*}[ht!]
\centering
\begin{tabularx}{\textwidth}{l|l|c|c|c|X|p{1cm}}
\hline
Model & \makecell{Training\\technique} & \makecell{Accuracy (\%)\\MNIST HWD.} & \makecell{No. trained conn.\\per input pin ($N_c$)} & \makecell{No. Gates\\/LUTs} & Remarks & Sections \\
\hline

\multirow{4}{*}{LGN}
& \makecell{Fixed-\\connections} & 97.92 & None & 48K & \vspace{-\baselineskip}
\begin{itemize}[label=$+$,leftmargin=10pt,itemindent=0pt,nosep]
    \item Proven to work well for deep networks
\end{itemize} 
\begin{itemize}[label=$-$,leftmargin=10pt,itemindent=0pt,nosep]
\item Training only the gates is suboptimal
\end{itemize}
& \ref{sec:standard_LGN_training}, \ref{sec:LGN_exp_partial}\\

\cline{2-7}
& \makecell{Partial-\\connections} & 98.83 & 8 & 24K & \vspace{-\baselineskip}
\begin{itemize}[label=$+$,leftmargin=10pt,itemindent=0pt,nosep]
    \item Significant accuracy increase compared to fixed-connection LGNs
    \item The number of trainable parameters increases only slightly
\end{itemize} 
\begin{itemize}[label=$-$,leftmargin=10pt,itemindent=0pt,nosep]
    \item Not all possible connections are taken into account
\end{itemize} 
& \ref{sec:LGNs_training_connections}, \ref{sec:LGN_exp_partial}\\
&  & 98.79 & 16 & 24K & & \\

\cline{2-7}
& \makecell{\textbf{All-}\\\textbf{connections}} & \textbf{98.92} & \textbf{all} & \textbf{16K} &\vspace{-\baselineskip} 
\begin{itemize}[label=$+$,leftmargin=10pt,itemindent=0pt,nosep]
    \item Highest accuracy achieved in this work
    \item State-of-the-art performance for an LGN up 16K gates or less
\end{itemize} 
\begin{itemize}[label=$-$,leftmargin=10pt,itemindent=0pt,nosep]
 \item Significantly more trainable parameters
\end{itemize} 
& \ref{sec:LGNs_training_connections}, \ref{sec:LGNs_all}\\

\hline
2-LUTN
& \makecell{Fixed-\\connections} & 98.09 & None & 48K &\vspace{-\baselineskip}
\begin{itemize}[label=$+$,leftmargin=10pt,itemindent=0pt,nosep]
    \item Slightly higher accuracy, shorter training times, and less trainable parameters compared to the LGNs
\end{itemize} 
\begin{itemize}[label=$-$,leftmargin=10pt,itemindent=0pt,nosep]
    \item Trainable connections have not yet been tested with 2-LUTNs
\end{itemize}
& \ref{sec:sigmoid_annealing}, \ref{sec:LUTs_vs_LGNs}\\

\hline
\multirow{4}{*}{6-LUTN}
& \makecell{Fixed-\\connections} & 98.15 & None & 6K & \vspace{-\baselineskip}
\begin{itemize}[label=$+$,leftmargin=10pt,itemindent=0pt,nosep]
    \item More tailored to FPGAs and requires less connections compared to LGNs
\end{itemize} 
\begin{itemize}[label=$-$,leftmargin=10pt,itemindent=0pt,nosep]
    \item Accuracy decreases for deep networks due to discretization of the LUT entries
    \item Training only the LUTs is suboptimal
\end{itemize}
& \ref{sec:sigmoid_annealing}, \ref{sec:2_LUTs_to_6_LUTs}\\

\cline{2-7}
& \makecell{Partial-\\connections} & 98.55 & 8 & 4K &\vspace{-\baselineskip}
\begin{itemize}[label=$+$,leftmargin=10pt,itemindent=0pt,nosep]
    \item Notable accuracy increase for shallow networks
\end{itemize} 
\begin{itemize}[label=$-$,leftmargin=10pt,itemindent=0pt,nosep]
    \item Accuracy becomes similar to LUTNs with fixed connections for deep networks
\end{itemize}
& \ref{sec:sigmoid_annealing}, \ref{sec:training_conn_LUTN}, \ref{sec:6_LUT_partial}\\
&  & 98.56 & 16 & 4K & & \\
\cline{2-7}
& \makecell{\textbf{All}-\\\textbf{connections}} & \textbf{98.88} & \textbf{all} & \textbf{4K }&\vspace{-\baselineskip}
\begin{itemize}[label=$+$,leftmargin=10pt,itemindent=0pt,nosep]
    \item Highest accuracy achieved with 6-LUTNs in this work
    \item State-of-the-art performance for a LUTN with 4K 6-LUTs or less
    \item Ideal for implementation on an FPGA
\end{itemize} 
\begin{itemize}[label=$-$,leftmargin=10pt,itemindent=0pt,nosep]
    \item Notable accuracy decrease for deep networks due to discretization
\end{itemize}
& \ref{sec:sigmoid_annealing}, \ref{sec:training_conn_LUTN}, \ref{sec:6_LUT_all}\\
\hline

\end{tabularx}
\caption{Overview of all different models and methods used in this work. For each model and method combination, the best accuracy (\%) on the MNIST Handwritten Digits data set is reported. Additionally the number of trained connections per gate or LUT input pin is given, alongside with the number of gates or LUTs in the whole network. Besides that, a list of advantages and disadvantages of each training method is presented, alongside the relevant sections for each method. Results indicated in bold are considered state-of-the-art, meaning that to our knowledge, these are the highest accuracies on the MNIST Handwritten Digits data set with a network of logic gates or LUTs of this size.}
\vspace{2cm}
\label{tab:overview}
\end{table*}

\subsection{Logic gate networks}
We introduced a method for partial and full training of the connections in deep differentiable logic gate networks (LGNs) and lookup table networks (LUTNs). When training the connections alongside the gates or LUTs, we maintain a similar accuracy while requiring significantly smaller networks compared to the fixed-connection counterparts. This improves the scalability of LGNs and LUTNs drastically, meaning that they are much better suited for AI applications on resource-constrained devices.\\

For the LGNs, several modifications to the standard algorithm from Section \ref{sec:standard_LGN_training} were necessary for stable training up to 10-layer deep networks. These fully trainable LGNs were benchmarked in Section \ref{sec:LGNs_all} on three different data sets: the Yin-Yang, MNIST Handwritten Digits and Fashion-MNIST data sets. To scale the training algorithm to such deep networks, we increased the learning rate as compared to fixed-connection LGNs that use a value of 0.01 \citep{petersen_deep_2022} to a value of 0.1, added straight-through estimators (STEs) \citep{kim_deep_2023} on the gates and connections, and removed constant-output gate types `0' and `1'. We saw most of the improvement using the higher learning rate. In addition, without the STE applied to the connections, there is no mechanism in place that makes sure that the model converges to just one connection per gate input pin. Instead, the model learns the best mixture of connections to decrease the loss. Introducing the STE on the connections, forces the model to compute the forward pass with the current optimal connections, but still takes into account the gradients of all other possible connections to each input pin in the backward pass. Another way of interpreting this is that the STE on the connections in the forward pass can be seen as training a fixed-connection LGN, only now we still take into account the other possible connections in the backward pass. The STE on the gates improved the results slightly. The removal of the constant-output gate types was beneficial, since these gate types prohibit information flow from one layer to the next one. Hence the removal of these gate types yielded the largest benefits for deeper networks. By employing these changes to the algorithm, we managed to achieve an accuracy of 98.45\% with one layer of 8000 gates on the MNIST Handwritten Digits data set. In comparison, the fixed-connection LGNs require 384000 gates to obtain 98.47\% \citep{petersen_deep_2022}. Hence, since we add our connection training method, we require almost 50 times fewer gates and connections while achieving a similar accuracy. In addition, we demonstrated that two layers of 8000 gates could achieve an accuracy of 98.92\%. To the best of our knowledge, this is the highest accuracy on the MNIST Handwritten Digits data set with up to 16000 gates. This result is given in Table \ref{tab:overview}, alongside the highest accuracy of all the other models and training techniques on the MNIST Handwritten Digits benchmark. This table shows that training the connections of LGNs increases the accuracy, using a noticeably smaller network. A fixed-connection LGN achieves 97.92\% using 48000 gates, whereas an LGN with fully trainable connections achieves 98.92\% with only 16000 gates. Continuing the discussion about the fully trainable LGNs, it was observed in Section \ref{sec:LGNs_all} that changing the width of the layers had a much more profound effect on the accuracy compared to changing the number of layers. Wider layers, generally speaking, lead to higher accuracies in networks where the total number of gates is kept constant. In most cases, the accuracy saturated for deep networks. It is not clear yet if this is because the LGNs are unable to learn more complex features in later layers, i.e. it is a fundamental problem, or if this is a limitation of the proposed training algorithm.\\

\subsection{Lookup table networks}
The LUT model and the training algorithm of the LUTNs were modified to enable stable training of deep networks. Since 6-LUTs are the fundamental computational building blocks of FPGAs, the LUTNs are able to deliver a near one-to-one mapping from a trained AI model to the FPGA fabric. This property makes them more attractive for low-power and low-latency AI on FPGAs. By putting forward a stable training method and introducing trainable connections, this work notably enhances their scalability, enabling more complex AI workloads to be mapped directly onto FPGA fabrics.\\

To realize these benefits, a new sigmoid-based annealing method was introduced in Section \ref{sec:sigmoid_annealing}. This method provided higher accuracies and shorter training times compared to the LGNs (Section \ref{sec:LUTs_vs_LGNs}) and the LUTN-training algorithm of \citet{mommen_inter-patient_2026} (Section \ref{sec:2_LUTs_to_6_LUTs}) on the MNIST Handwritten Digits data set. In addition, the LUTNs consisting of 2-input LUTs, required less trainable parameters compared to the LGNs on the same benchmark (Section \ref{sec:LUTs_vs_LGNs}), thus offering a simpler description of the same underlying hardware model. Upon increasing the number of LUT input pins to $N=6$ in Section \ref{sec:2_LUTs_to_6_LUTs}, we showed that the accuracy increased, since each neuron is capable of learning $2^{N^N}=1.845\cdot10^{19}$ binary functions instead of only 16. The proposed sigmoid-based annealing method, Section \ref{sec:sigmoid_annealing}, showed a decrease in training time up to a factor of 13.62 compared to \citet{mommen_inter-patient_2026} for a 5-layer 6-LUTN in Section \ref{sec:2_LUTs_to_6_LUTs}. The networks of \citet{bacellar_differentiable_2024} do not mention any training times, and therefore, can not be compared with. The only remaining drawback of the sigmoid-based annealing method is the emerging discretization gap for deep networks consisting of 6-LUTs. Before annealing, the accuracy difference between the full precision model and the binarized model is at its largest. During binarization (i.e. sigmoid-based annealing), this difference decreases, mainly because the performance of the full precision model decreases. The gap exists, since the LUT entries do not converge to binary values on their own, hence why the sigmoid annealing is performed. An investigation of gradient dynamics for 6-LUTNs will be necessary to pinpoint why this gap emerged and how it can be dealt with. A different binarization method might be needed or not all LUT entries should be binarized at the same time. Note that the non-binarized network cannot be implemented in hardware, and is reported to assess how well our training method works.\\

The algorithm for training connections introduced in Section \ref{sec:training_conn_LUTN} was evaluated on the LUTNs. Both for partial training (Section \ref{sec:6_LUT_partial}) and full training (Section \ref{sec:6_LUT_all}) of the connections, the LUTNs outperformed the fixed-connection LUTNs on the MNIST Handwritten Digits data set up to 6-layer deep networks. We observed that without binarizing the LUTs, the fully trainable LUTNs easily get above 99\% on this data set, e.g. two layers of 2000 6-LUTs achieve 99.1\%. Again, the only drawback of this model is the training of the LUT entries. As we binarize the LUT entries, the accuracy decreases. It is important to note that the network with full precision LUT entries is not implementable on hardware, and is purely used to compare to the fully binarized network to help determine the shortcomings of the proposed training algorithm. An investigation of the gradient dynamics is needed to determine why this discretization gap occurs for deep 6-LUTNs. The performance of the 2-LUTNs and 6-LUTNs on the MNIST Handwritten Digits data set is shown in Table \ref{tab:overview}. Similarly to the LGNs, making the connections trainable increases the performance substantially. A fixed-connection 6-LUTN requires 6000 LUTs to attain 98.15\% accuracy, while a 6-LUTN with fully trainable connections achieves 98.88\% accuracy using only 4000 LUTs. To our knowledge, this is the highest accuracy achieved on the MNIST Handwritten Digits benchmark for 4000 or fewer 6-LUTs. The Differentiable Weightless Neural Network of \citet{bacellar_differentiable_2024} achieved a lower accuracy of 98.77\% with a larger model consisting of 4.6K 6-LUTs.

\subsection{Outlook and future work}
\begin{table*}[t]
\centering
\begin{tabularx}{\textwidth}{X|X|X|X}
\hline
\textbf{Research direction} &
\textbf{Type of work} &
\textbf{Current status} &
\textbf{Expected benefit} \\
\hline

Connection-optimized LGNs &
Extension of existing LGN architectures &
Convolutional and recurrent LGNs have already been realized \citep{petersen_convolutional_2024,buhrer_recurrent_2025}, but their connections remain fixed. &
Noticeable reductions in gate count while maintaining a similar performance. \\
\hline

Convolutional LUTNs &
New architecture &
Not yet realized. &
FPGA-friendly image and video processing with reduced model size. \\
\hline

Recurrent LUTNs &
New architecture &
Not yet realized. &
FPGA-friendly sequence processing with reduced model size. \\
\hline

Deeper LGNs and LUTNs &
Training methodology &
Stable training demonstrated up to 10-layer LGNs and 6-layer LUTNs. &
Improved scalability and training stability in deeper networks.\\
\hline

Alternative training methods &
Training methodology &
Current work focuses primarily on differentiable training with backpropagation. &
More tailored training method for inherently non-differentiable networks might perform better. \\
\hline
\end{tabularx}
\caption{The potential directions for future work. The most straightforward directions include applying the connection training algorithm to already existing architectures with fixed connections, and improving the algorithm for deeper networks. Another direction is the development of convolutional and recurrent LUTNs, which are especially beneficial for low-latency and low-power AI applications on FPGAs. An alternative training method without differentiable logic gates, LUTs or connections, that takes the discrete nature of the LGNs and LUTNs into account, might be more suitable for these types of networks.}
\label{tab:future_work}
\end{table*}

This work only mentioned feedforward architectures of both the LGNs and LUTNs. Convolutional and recurrent architectures have already been realized for the LGNs. Therefore, applying the connection training algorithm to convolutional and recurrent architectures could significantly shrink the total number of gates for those architectures, while maintaining a similar accuracy. For example, to achieve 86.29\% on the CIFAR10 data set, a convolutional LGN consisting of 61 million gates was necessary \citep{petersen_convolutional_2024}. In addition, to achieve a BLEU score of 5.00 on the WMT’14 English-German translation task, around 1.5 million gates were needed \citep{buhrer_recurrent_2025}. In both cases, training the connections would most likely decrease the gate count considerably. This research direction can be found in Table \ref{tab:future_work}\\

Next to applying the connection training algorithm to existing architectures, we could also first come up with novel architectures, and next apply the connection training algorithm to decrease the model size. One example is convolutional lookup table networks, which, at the time of writing have not yet been realized. Here, one directly trains a convolutional network consisting entirely out of lookup tables. The LUTNs do not require any conversion step to be implemented in hardware, since they consist entirely out of LUTs. These convolutional LUTNs would be more suited for image or video related tasks on FPGAs. The LUTNs use the actual computational elements of an FPGA, namely the 6-LUT, which is not the case for the LGNs. Hence, it is expected that the convolutional LUTNs would be a more suitable model compared to convolutional LGNs on FPGAs. We can make the same case for recurrent LUTNs. These have not been realized yet and are expected to be more suited for tasks on FPGAs than the recurrent LGNs. Both research directions can be found in Table \ref{tab:future_work}.\\

Lastly, the connection training algorithm that we introduced in Section \ref{sec:LGNs_training_connections} and Section \ref{sec:training_conn_LUTN} could be improved with the goal of training networks with more than 10 layers with shorter training times and for which the accuracy increases with network depth. In this work, we only tested the training stability up to 10 layers. However, networks topologies such as convolutional networks and LLMs require significantly more layers. It remains to be seen if the connection training algorithm will still operate in a stable manner for such deep networks. Possibly, alternative training methods should be researched (Table \ref{tab:future_work}). The training method employed in this work relies on backpropagation. However, LGNs and LUTNs are inherently discrete-valued, meaning that relaxing the components of the networks to differentiable expressions might be disadvantageous for their performance. Instead, evolutionary optimization  could be a more suitable replacement for backpropagation \citep{salimans_evolution_2017,such_deep_2018,sarkar_evolution_2026}.

\section*{CRediT authorship contribution statement}
\textbf{Wout Mommen:} Conceptualization, Methodology, Software, Validation, Formal analysis, Investigation, Writing -- Original Draft, Visualization.\\
\textbf{Lars Keuninckx:} Conceptualization, Methodology, Validation, Formal analysis, Writing -- Original Draft, Supervision.\\
\textbf{Matthias Hartmann:} Resources, Project administration, Funding acquisition.\\
\textbf{Werner Van Leekwijck:} Resources, Project administration, Funding acquisition.\\
\textbf{Piet Wambacq:} Resources, Supervision, Project administration, Funding acquisition.\\

\section*{Declaration of competing interest}
The authors declare that they have no known competing financial interests or personal relationships that could have appeared to influence the work reported in this paper. 

\section*{Acknowledgment}
This work has received funding from the Flemish Government under ``Onderzoeksprogramma Artificiële Intelligentie Vlaanderen'' (Flanders AI Research).

\appendix
\section{Derivations and proofs} \label{sec:appendix}
\subsection{Derivation of the LUT-neuron model} \label{sec:Derivation_LUT_Model}
In Section \ref{sec:sigmoid_annealing}, we described the LUT model as the Boolean equation of a MUX. In addition, we added sigmoid-based annealing to have a binarization mechanism and keep the LUT-entries between zero and one during training. To show that the output value of a LUT \eqref{eq:LUT_eq} is equal to the expectation value of the LUT entries \eqref{eq:LUT_eq_3}, we must proof the intermediate step \eqref{eq:LUT_eq_2}. We need to show that $p_i=\prod_{j=0}^{N-1} \overline{s_{ij} L_j} + s_{ij} L_j$ indeed is a probability distribution. This means that
\begin{align}
    \sum_{i=0}^{2^N-1}p_i &= 1, \label{eq:sum_prob}\\
    \text{and } \forall i:p_i &\in[0,1]. \label{eq:prob}
\end{align}

\noindent
\textbf{Proof by induction of \eqref{eq:sum_prob}.}\\

For $N=1$ we have that 
\begin{equation*}
    \overline{L_0} + L_0 = (1-L_0) + L_0 = 1.
\end{equation*}
Note that here $\mathbf{s}$ is a $2^N\times N$ matrix listing all possible $N$-bit numbers, so in this case $\mathbf{s}= \begin{pmatrix}
    0 \\
    1
\end{pmatrix}$.

Assume the equation holds for $N=n$:\\
\begin{align*}
    &\sum_{i=0}^{2^n-1}\prod_{j=0}^{n-1} \left(\overline{s_{ij}L_j} + s_{ij}L_j\right) \\
    &=\overline{L_0L_1\cdots L_{n-2} L_{n-1}}  + \overline{L_0L_1\cdots L_{n-2}}L_{n-1} + \cdots \\
    &+ L_0L_1\cdots L_{n-2}L_{n-1} \label{eq:induction_n}\\
    &=1.
\end{align*}

Now we prove that the equation holds for $N=n+1$ given the above expression for $N=n$:
\begin{align*}
    &(\overline{L_0L_1\cdots L_{n-2} L_{n-1}})\,\overline{L_n}  + (\overline{L_0L_1\cdots L_{n-2}}L_{n-1}) \overline{L_n} + \cdots \\ 
    &+ (L_0L_1\cdots L_{n-2}L_{n-1})\overline{L_n} \\
    &+ (\overline{L_0L_1\cdots L_{n-2} L_{n-1}})L_n  + (\overline{L_0L_1\cdots L_{n-2}}L_{n-1}) L_n + \cdots \\ 
    &+ (L_0L_1\cdots L_{n-2}L_{n-1}) L_n \\
    \\
    &=\big(\overline{L_0L_1\cdots L_{n-2} L_{n-1}} + \overline{L_0L_1\cdots L_{n-2}}L_{n-1} + \cdots \\ 
    &+ L_0L_1\cdots L_{n-2}L_{n-1}\big)\overline{L_n} \\
    &+\big(\overline{L_0L_1\cdots L_{n-2} L_{n-1}} + \overline{L_0L_1\cdots L_{n-2}}L_{n-1} + \cdots \\ 
    &+ L_0L_1\cdots L_{n-2}L_{n-1}\big)L_n \\
    \\
    &= 1 \cdot (1-L_n) + 1 \cdot L_n \\
    &= 1.\,\square\\
\end{align*}

\noindent
\textbf{Proof of \eqref{eq:prob}}
Since each $p_i$ is a product of combinations of the input values $L_j$ or the inverse thereof (i.e. $\overline{L_j}$), it holds that $\forall j: L_j, \overline{L_j} \in [0,1] \Rightarrow \forall i: p_i \in [0,1].\,\square$

\subsection{Derivation of GEMM-implementation of LUT neuron model} \label{sec:Derivation_GEMM}
In Section \ref{sec:sigmoid_annealing}, we showed that the equation of the proposed LUT model containing element-wise operations can also be written employing much faster matrix multiplications. To proof the equations \eqref{eq:LUT_GEMMs} and \eqref{eq:LUT_GEMMs_p}, we first need to rewrite $L_{out,b,k}$ as follows:
\begin{align*}
    L_{out,b,k} &= \sum_{i=0}^{2^N-1} \sigma(\beta W_{i,k}) \prod_{j=0}^{N-1}(\bar{s}_{i,j} \bar{L}_{b,k,j} + s_{i,j} L_{b,k,j})\\
    &= \sum_{i=0}^{2^N-1} \sigma(\beta W_{i,k}) p_{b,k,i}\\
    &= \sum_{i=0}^{2^N-1} \exp[\log(\sigma(\beta W_{i,k})) + \log(p_{b,k,i})]\\
    &= \exp{\operatorname{LSE}_i[\log(\sigma(\beta W_{i,k})) + \log(p_{b,k,i})]}.
\end{align*}
Here $LSE$ is the LogSumExp function, which is defined as $\operatorname{LSE}(x_1,\dots,x_M)\equiv\operatorname{log}(\sum_i\operatorname{exp}(x_i))$. Noting that $s_{i,j}$ is binary:
\begin{align*}
    \log(p_{b,k,i}) &= \log\left(\prod_{j=0}^{N-1}(\bar{s}_{i,j} \bar{L}_{b,k,j} + s_{i,j} L_{b,k,j})\right)\\
    &= \sum_{j=0}^{N-1}\log(\bar{s}_{i,j} \bar{L}_{b,k,j} + s_{i,j} L_{b,k,j})\\
    &= \sum_{j=0}^{N-1}\bar{s}_{i,j} \log(\bar{L}_{b,k,j}) + s_{i,j} \log(L_{b,k,j})\\
    &= \sum_{j=0}^{2N-1} \tilde{s}_{i,j} \operatorname{log}(\tilde{L}_{b,k,j})\\
    &= \sum_{j=0}^{2N-1} \operatorname{log}(\tilde{L}_{b,k,j}) (\tilde{\mathbf{s}}^T)_{j,i}\\
    &= \left[\operatorname{log}(\tilde{\mathbf{L}}_{b,k}) \tilde{\mathbf{s}}^T\right]_i.
\end{align*}
Note that PyTorch computes this last expression as a batched matrix multiplication with dimensions $(B,K,2N)\times(2N,2^N)$, where $B$ is the batch size, $K$ is the layer size, $N$ are the number of LUT input pins and $2^N$ are the number of LUT entries.

\bibliographystyle{elsarticle-harv} 
\bibliography{refs}

@misc{bacellar_differentiable_2024,
  title = {Differentiable {{Weightless Neural Networks}}},
  author = {Bacellar, Alan T. L. and Susskind, Zachary and Jr, Mauricio Breternitz and John, Eugene and John, Lizy K. and Lima, Priscila M. V. and Fran{\c c}a, Felipe M. G.},
  year = 2024,
  month = nov,
  number = {arXiv:2410.11112},
  publisher = {arXiv},
  doi = {10.48550/arXiv.2410.11112},
  urldate = {2024-11-22},
  archiveprefix = {arXiv}
}

@article{desislavov_trends_2023,
  title = {Trends in {{AI}} Inference Energy Consumption: {{Beyond}} the Performance-vs-Parameter Laws of Deep Learning},
  shorttitle = {Trends in {{AI}} Inference Energy Consumption},
  author = {Desislavov, Radosvet and {Mart{\'i}nez-Plumed}, Fernando and {Hern{\'a}ndez-Orallo}, Jos{\'e}},
  year = 2023,
  month = apr,
  journal = {Sustainable Computing: Informatics and Systems},
  volume = {38},
  pages = {100857},
  issn = {2210-5379},
  doi = {10.1016/j.suscom.2023.100857},
  urldate = {2025-04-03}
}

@inproceedings{diehl_truehappiness_2016,
  title = {{{TrueHappiness}}: {{Neuromorphic}} Emotion Recognition on {{TrueNorth}}},
  shorttitle = {{{TrueHappiness}}},
  booktitle = {2016 {{International Joint Conference}} on {{Neural Networks}} ({{IJCNN}})},
  author = {Diehl, Peter U. and Pedroni, Bruno U. and Cassidy, Andrew and Merolla, Paul and Neftci, Emre and Zarrella, Guido},
  year = 2016,
  month = jul,
  pages = {4278--4285},
  doi = {10.1109/IJCNN.2016.7727758},
  urldate = {2025-04-06}
}

@article{eshraghian_training_2023,
  title = {Training {{Spiking Neural Networks Using Lessons From Deep Learning}}},
  author = {Eshraghian, Jason K. and Ward, Max and Neftci, Emre O. and Wang, Xinxin and Lenz, Gregor and Dwivedi, Girish and Bennamoun, Mohammed and Jeong, Doo Seok and Lu, Wei D.},
  year = 2023,
  month = sep,
  journal = {Proceedings of the IEEE},
  volume = {111},
  number = {9},
  pages = {1016--1054},
  doi = {10.1109/JPROC.2023.3308088},
  urldate = {2025-04-03}
}

@article{hubara_quantized_2018,
  title = {Quantized {{Neural Networks}}: {{Training Neural Networks}} with {{Low Precision Weights}} and {{Activations}}},
  shorttitle = {Quantized {{Neural Networks}}},
  author = {Hubara, Itay and Courbariaux, Matthieu and Soudry, Daniel and {El-Yaniv}, Ran and Bengio, Yoshua},
  year = 2018,
  journal = {Journal of Machine Learning Research},
  volume = {18},
  number = {187},
  pages = {1--30},
  urldate = {2025-04-03}
}

@article{kim_spikingyolo_2020,
  title = {Spiking-{{YOLO}}: {{Spiking Neural Network}} for {{Energy-Efficient Object Detection}}},
  shorttitle = {Spiking-{{YOLO}}},
  author = {Kim, Seijoon and Park, Seongsik and Na, Byunggook and Yoon, Sungroh},
  year = 2020,
  month = apr,
  journal = {Proceedings of the AAAI Conference on Artificial Intelligence},
  volume = {34},
  number = {07},
  pages = {11270--11277},
  doi = {10.1609/aaai.v34i07.6787},
  urldate = {2025-04-06},
  copyright = {Copyright (c) 2020 Association for the Advancement of Artificial Intelligence},
  langid = {english}
}

@misc{kriener_yinyang_2022,
  title = {The {{Yin-Yang}} Dataset},
  author = {Kriener, Laura and G{\"o}ltz, Julian and Petrovici, Mihai A.},
  year = 2022,
  month = jan,
  number = {arXiv:2102.08211},
  publisher = {arXiv},
  doi = {10.48550/arXiv.2102.08211},
  urldate = {2025-04-04},
  archiveprefix = {arXiv}
}

@misc{lecun_mnist_1998,
  title = {MNIST Database of Handwritten Digits},
  author = {LeCun, Yann and Cortes, Corinna and Burges, Christopher J. C.},
  year = {1998},
  howpublished = {\url{http://yann.lecun.com/exdb/mnist/}},
  urldate = {2025-04-04}
}

@article{petersen_convolutional_2024,
  title = {Convolutional {{Differentiable Logic Gate Networks}}},
  author = {Petersen, Felix and Kuehne, Hilde and Borgelt, Christian and Welzel, Julian and Ermon, Stefano},
  year = 2024,
  month = dec,
  journal = {Advances in Neural Information Processing Systems},
  volume = {37},
  pages = {121185--121203},
  urldate = {2025-04-03},
  langid = {english}
}

@article{petersen_deep_2022,
  title = {Deep {{Differentiable Logic Gate Networks}}},
  author = {Petersen, Felix and Borgelt, Christian and Kuehne, Hilde and Deussen, Oliver},
  year = 2022,
  month = dec,
  journal = {Advances in Neural Information Processing Systems},
  volume = {35},
  pages = {2006--2018},
  urldate = {2025-04-03},
  langid = {english}
}

@inproceedings{putra_qspinn_2021,
  title = {Q-{{SpiNN}}: {{A Framework}} for {{Quantizing Spiking Neural Networks}}},
  shorttitle = {Q-{{SpiNN}}},
  booktitle = {2021 {{International Joint Conference}} on {{Neural Networks}} ({{IJCNN}})},
  author = {Putra, Rachmad Vidya Wicaksana and Shafique, Muhammad},
  year = 2021,
  month = jul,
  pages = {1--8},
  doi = {10.1109/IJCNN52387.2021.9534087},
  urldate = {2025-04-03}
}

@inproceedings{rastegari_xnornet_2016,
  title = {{{XNOR-Net}}: {{ImageNet Classification Using Binary Convolutional Neural Networks}}},
  shorttitle = {{{XNOR-Net}}},
  booktitle = {Computer {{Vision}} -- {{ECCV}} 2016},
  author = {Rastegari, Mohammad and Ordonez, Vicente and Redmon, Joseph and Farhadi, Ali},
  editor = {Leibe, Bastian and Matas, Jiri and Sebe, Nicu and Welling, Max},
  year = 2016,
  pages = {525--542},
  publisher = {Springer International Publishing},
  address = {Cham},
  doi = {10.1007/978-3-319-46493-0_32},
  langid = {english}
}

@article{rathi_stdpbased_2019,
  title = {{{STDP-Based Pruning}} of {{Connections}} and {{Weight Quantization}} in {{Spiking Neural Networks}} for {{Energy-Efficient Recognition}}},
  author = {Rathi, Nitin and Panda, Priyadarshini and Roy, Kaushik},
  year = 2019,
  month = apr,
  journal = {IEEE Transactions on Computer-Aided Design of Integrated Circuits and Systems},
  volume = {38},
  number = {4},
  pages = {668--677},
  doi = {10.1109/TCAD.2018.2819366},
  urldate = {2025-04-03}
}

@article{schuman_opportunities_2022,
  title = {Opportunities for Neuromorphic Computing Algorithms and Applications},
  author = {Schuman, Catherine D. and Kulkarni, Shruti R. and Parsa, Maryam and Mitchell, J. Parker and Date, Prasanna and Kay, Bill},
  year = 2022,
  month = jan,
  journal = {Nature Computational Science},
  volume = {2},
  number = {1},
  pages = {10--19},
  doi = {10.1038/s43588-021-00184-y},
  copyright = {2022 Springer Nature America, Inc.}
}

@inproceedings{sevilla_compute_2022,
  title = {Compute {{Trends Across Three Eras}} of {{Machine Learning}}},
  booktitle = {2022 {{International Joint Conference}} on {{Neural Networks}} ({{IJCNN}})},
  author = {Sevilla, Jaime and Heim, Lennart and Ho, Anson and Besiroglu, Tamay and Hobbhahn, Marius and Villalobos, Pablo},
  year = 2022,
  month = jul,
  pages = {1--8},
  doi = {10.1109/IJCNN55064.2022.9891914}
}

@misc{xiao_fashionmnist_2017,
  title = {Fashion-{{MNIST}}: A {{Novel Image Dataset}} for {{Benchmarking Machine Learning Algorithms}}},
  shorttitle = {Fashion-{{MNIST}}},
  author = {Xiao, Han and Rasul, Kashif and Vollgraf, Roland},
  year = 2017,
  month = sep,
  number = {arXiv:1708.07747},
  publisher = {arXiv},
  doi = {10.48550/arXiv.1708.07747},
  urldate = {2025-04-04},
  archiveprefix = {arXiv}
}

@article{kim_deep_2023,
	title = {Deep {Stochastic} {Logic} {Gate} {Networks}},
	volume = {11},
	copyright = {https://creativecommons.org/licenses/by-nc-nd/4.0/},
	doi = {10.1109/ACCESS.2023.3328622},
	urldate = {2024-05-07},
	journal = {IEEE Access},
	author = {Kim, Youngsung},
	year = {2023},
	pages = {122488--122501},
}

@misc{mommen_inter-patient_2026,
	title = {Inter-patient {ECG} {Arrhythmia} {Classification} with {LGNs} and {LUTNs}},
	doi = {10.48550/arXiv.2601.11433},
	urldate = {2026-01-29},
	publisher = {arXiv},
	author = {Mommen, Wout and Keuninckx, Lars and Detterer, Paul and Colpaert, Achiel and Wambacq, Piet},
	month = jan,
	year = {2026},
}

@misc{buhrer_recurrent_2025,
	title = {Recurrent {Deep} {Differentiable} {Logic} {Gate} {Networks}},
	doi = {10.48550/arXiv.2508.06097},
	urldate = {2025-09-11},
	publisher = {arXiv},
	author = {Bührer, Simon and Plesner, Andreas and Aczel, Till and Wattenhofer, Roger},
	month = aug,
	year = {2025},
}

@misc{gerlach_warp-luts_2025,
	title = {{WARP}-{LUTs} - {Walsh}-{Assisted} {Relaxation} for {Probabilistic} {Look} {Up} {Tables}},
	doi = {10.48550/arXiv.2510.15655},
	urldate = {2025-11-06},
	publisher = {arXiv},
	author = {Gerlach, Lino and Våge, Liv and Gerlach, Thore and Kauffman, Elliott},
	month = oct,
	year = {2025},
}

@inproceedings{umuroglu_logicnets_2020,
	title = {{LogicNets}: {Co}-{Designed} {Neural} {Networks} and {Circuits} for {Extreme}-{Throughput} {Applications}},
	shorttitle = {{LogicNets}},
	doi = {10.1109/FPL50879.2020.00055},
	urldate = {2025-12-09},
	booktitle = {2020 30th {International} {Conference} on {Field}-{Programmable} {Logic} and {Applications} ({FPL})},
	author = {Umuroglu, Yaman and Akhauri, Yash and Fraser, Nicholas James and Blott, Michaela},
	month = aug,
	year = {2020},
	pages = {291--297},
}

@inproceedings{andronic_polylut_2023,
	title = {{PolyLUT}: {Learning} {Piecewise} {Polynomials} for {Ultra}-{Low} {Latency} {FPGA} {LUT}-based {Inference}},
	shorttitle = {{PolyLUT}},
	doi = {10.1109/ICFPT59805.2023.00012},
	urldate = {2025-12-09},
	booktitle = {2023 {International} {Conference} on {Field} {Programmable} {Technology} ({ICFPT})},
	author = {Andronic, Marta and Constantinides, George A.},
	month = dec,
	year = {2023},
	pages = {60--68},
}

@inproceedings{andronic_neuralut_2024,
	title = {{NeuraLUT}: {Hiding} {Neural} {Network} {Density} in {Boolean} {Synthesizable} {Functions}},
	shorttitle = {{NeuraLUT}},
	doi = {10.1109/FPL64840.2024.00028},
	urldate = {2025-12-09},
	booktitle = {2024 34th {International} {Conference} on {Field}-{Programmable} {Logic} and {Applications} ({FPL})},
	author = {Andronic, Marta and Constantinides, George A.},
	month = sep,
	year = {2024},
	pages = {140--148},
}

@inproceedings{nazemi_energy-efficient_2019,
	title = {Energy-{Efficient}, {Low}-{Latency} {Realization} of {Neural} {Networks} {Through} {Boolean} {Logic} {Minimization}},
	doi = {10.1145/3287624.3287722},
	urldate = {2025-12-15},
	booktitle = {2019 24th {Asia} and {South} {Pacific} {Design} {Automation} {Conference} ({ASP}-{DAC})},
	author = {Nazemi, Mahdi and Pasandi, Ghasem and Pedram, Massoud},
	month = jan,
	year = {2019},
	pages = {1--6},
}

@article{keuninckx_training_2026,
	title = {On training networks of monostable multivibrator timer neurons},
	volume = {194},
	doi = {10.1016/j.neunet.2025.108092},
	urldate = {2026-02-05},
	journal = {Neural Networks},
	author = {Keuninckx, Lars and Hartmann, Matthias and Detterer, Paul and Safa, Ali and Mommen, Wout and Ocket, Ilja},
	month = feb,
	year = {2026},
	pages = {108092},
}

@misc{salimans_evolution_2017,
	title = {Evolution {Strategies} as a {Scalable} {Alternative} to {Reinforcement} {Learning}},
	doi = {10.48550/arXiv.1703.03864},
	urldate = {2026-07-09},
	publisher = {arXiv},
	author = {Salimans, Tim and Ho, Jonathan and Chen, Xi and Sidor, Szymon and Sutskever, Ilya},
	month = sep,
	year = {2017},
}

@misc{such_deep_2018,
	title = {Deep {Neuroevolution}: {Genetic} {Algorithms} {Are} a {Competitive} {Alternative} for {Training} {Deep} {Neural} {Networks} for {Reinforcement} {Learning}},
	shorttitle = {Deep {Neuroevolution}},
	doi = {10.48550/arXiv.1712.06567},
	urldate = {2026-07-09},
	publisher = {arXiv},
	author = {Such, Felipe Petroski and Madhavan, Vashisht and Conti, Edoardo and Lehman, Joel and Stanley, Kenneth O. and Clune, Jeff},
	month = apr,
	year = {2018},
}

@misc{sarkar_evolution_2026,
	title = {Evolution {Strategies} at the {Hyperscale}},
	doi = {10.48550/arXiv.2511.16652},
	urldate = {2026-07-09},
	publisher = {arXiv},
	author = {Sarkar, Bidipta and Fellows, Mattie and Duque, Juan Agustin and Letcher, Alistair and Villares, Antonio León and Sims, Anya and Wibault, Clarisse and Samsonov, Dmitry and Cope, Dylan and Liesen, Jarek and Li, Kang and Seier, Lukas and Wolf, Theo and Berdica, Uljad and Mohl, Valentin and Goldie, Alexander David and Courville, Aaron and Sevegnani, Karin and Whiteson, Shimon and Foerster, Jakob Nicolaus},
	month = feb,
	year = {2026},
}

\end{document}